\documentclass{opendatalab}

\usepackage[utf8]{inputenc}
\usepackage[table]{xcolor}
\usepackage[most]{tcolorbox}
\usepackage{longtable}
\usepackage{listings}
\usepackage[numbers]{natbib}
\usepackage{enumitem}
\usepackage{graphicx}
\usepackage{xspace} 
\usepackage{hyperref}
\usepackage{multirow}
\usepackage{caption}

\definecolor{codegreen}{rgb}{0,0.6,0}
\definecolor{codegray}{rgb}{0.5,0.5,0.5}
\definecolor{codepurple}{rgb}{0.58,0,0.82}
\definecolor{backcolour}{rgb}{0.95,0.95,0.92}
\definecolor{promptcolor}{HTML}{D1D0F2}
\definecolor{promptcolorheader}{HTML}{bdbcec}

\newcommand{\github}{\raisebox{-1.5pt}{\includegraphics[height=1.05em]{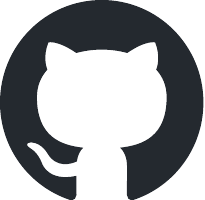}}\xspace}

\newcommand{\huggingface}{\raisebox{-1.5pt}{\includegraphics[height=1.05em]{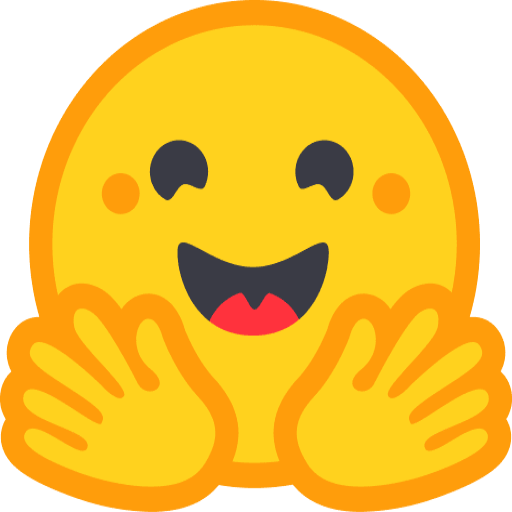}}\xspace}

\definecolor{promptcolor}{HTML}{D1D0F2}
\definecolor{promptcolorheader}{HTML}{bdbcec}

\newtcolorbox{promptbox}[1][]{
  enhanced, breakable,
  top=0.3em,bottom=0.3em,left=0.5em,right=0.5em,
  toptitle=0.3em,bottomtitle=0.2em,boxsep=0pt,
  colframe=promptcolorheader, colback=promptcolor!50, boxrule=0.5pt,
  width=\columnwidth, 
  title={\footnotesize #1} 
}
\lstdefinestyle{promptstyle}{
    backgroundcolor=\color{backcolour},   
    commentstyle=\color{codegreen},
    keywordstyle=\color{magenta},
    numberstyle=\tiny\color{codegray},
    stringstyle=\color{codepurple},
    basicstyle=\ttfamily\footnotesize,
    breakatwhitespace=false,         
    breaklines=true,                 
    captionpos=b,                    
    keepspaces=true,                 
    numbers=left,                    
    numbersep=5pt,                  
    showspaces=false,                
    showstringspaces=false,
    showtabs=false,                  
    tabsize=2
}
\definecolor{iris}{HTML}{ffffff}
\definecolor{white}{HTML}{dadde2}
\definecolor{chisa}{HTML}{866f83}

\title{Decouple to Generalize: Context-First Self-Evolving Learning for Data-Scarce Vision-Language Reasoning}

\author[1,2]{Tingyu Li}
\author[3]{Zheng Sun}
\author[3]{Jingxuan Wei}
\author[4]{Siyuan Li}
\author[1]{Conghui He}
\author[1]{Lijun Wu}
\author[1]{Cheng Tan}

\affiliation[1]{Shanghai Artificial Intelligence Laboratory}
\affiliation[2]{Shanghai JiaoTong University}
\affiliation[3]{University of Chinese Academy of Sciences}
\affiliation[4]{Zhejiang University}

\abstract{
Recent vision-language models (VLMs) achieve remarkable reasoning through reinforcement learning (RL), which provides a feasible solution for realizing continuous self-evolving large vision-language models (LVLMs) in the era of experience. However, RL for VLMs requires abundant high-quality multimodal data—especially challenging in specialized domains like chemistry, earth sciences, and multimodal mathematics. Existing strategies such as synthetic data and self-rewarding mechanisms suffer from limited distributions and alignment difficulties, ultimately causing reward hacking: models exploit high-reward patterns, collapsing policy entropy and destabilizing training. We propose \textbf{DoGe} (Decouple to Generalize), a dual-decoupling framework that guides models to first learn from context rather than problem solving by refocusing on the problem context scenarios overlooked by synthetic data methods. By decoupling learning process into dual components (Thinker and Solver), we reasonably quantify the reward signals of this process and propose a two-stage RL post-training approach from freely exploring context to practically solving tasks. Second, to increase the diversity of training data, DoGe constructs an evolving curriculum learning pipeline: an expanded native domain knowledge corpus and an iteratively evolving seed problems pool. Experiments show that our method consistently outperforms the baseline across various benchmarks, providing a scalable pathway for realizing self-evolving LVLMs.
}

\date{\today}
\correspondence{Cheng Tan, \email{tancheng@pjlab.org.cn}
\\
\\
  \parbox{\linewidth}{\centering
    \github~\href{https://github.com/orgs/opendatalab-raiser/DoGe}{\textbf{Code}} \quad
    \huggingface~\href{https://huggingface.co/datasets/opendatalab-raiser/DoGe}{\textbf{Dataset}}
  }
}

\begin{document}

\maketitle


\begin{figure}[ht]
\centering
\includegraphics[width=1.0\linewidth]{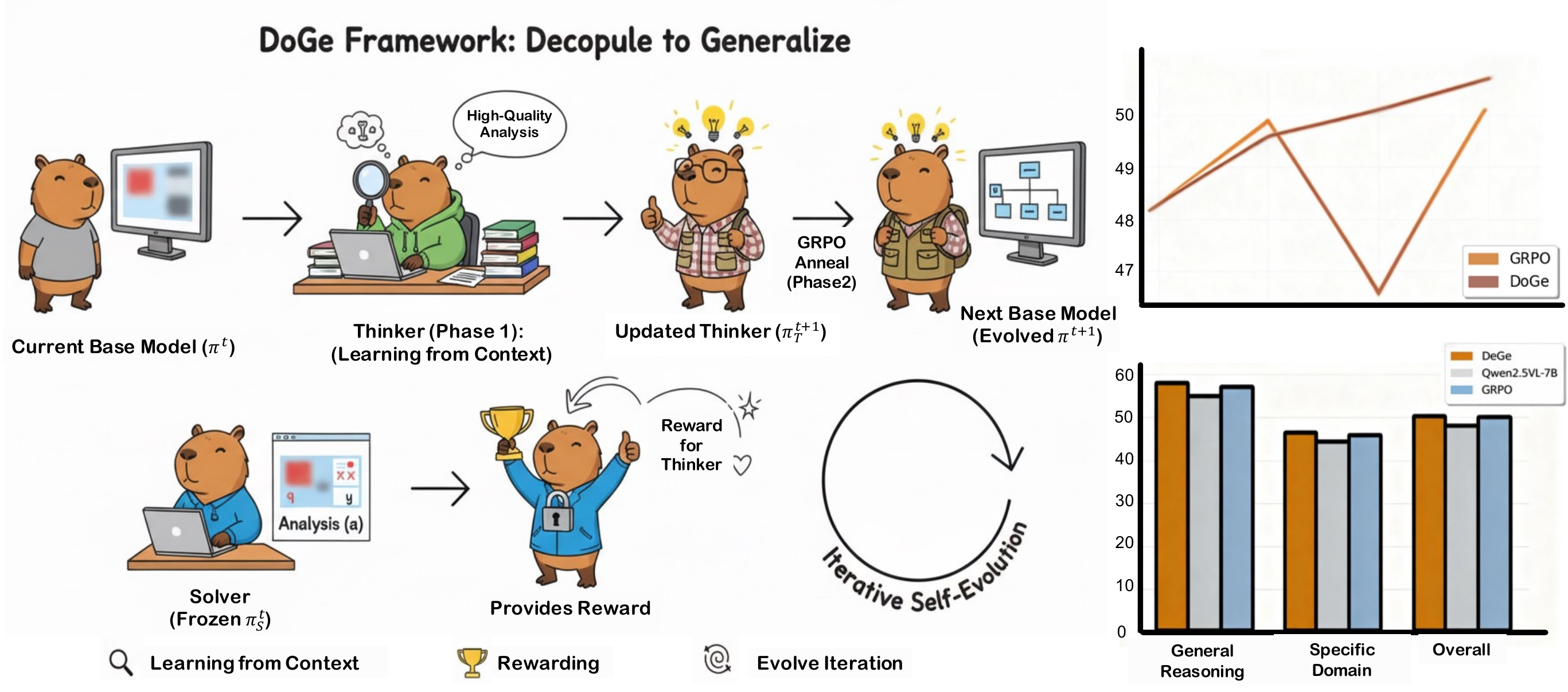} 
\caption{DoGe decouples the self-evolving VLM cognitive process into a "learning-application" cycle, designing two components: a learnable Thinker and a frozen Solver. In the first stage, the pass rate of the Solver is used as the quantitative reward for the Analysis (the output of the Thinker). In the second stage, standard GRPO is implemented for annealing, forming a complete iterative closed loop.}
\label{fig:intro}
\end{figure}

\section{Introduction}
\label{sec:intro}

Recent advances in VLMs have unlocked remarkable reasoning capabilities, moving beyond simple perception tasks to engage with complex, multi-step problems. By integrating RL post-training techniques, models can generate extended chain of thoughts, providing better responses for multi-modal queries~\cite{huang2025vision,yang2025qwen3}. This offers a viable solution for implementing self-evolving Agents in the era of experience~\cite{silver2025welcome}. However, the efficacy of these training paradigms is fundamentally tethered to the availability of abundant, high-quality multi-modal training data. For specialized, high-value domains such as chemistry, earth science, and even multi-modal mathematics, significant costs are often required to manually annotate domain-specific multi-modal knowledge corpora and design high-quality reasoning training data~\cite{parashar2024neglected,yarom2023you}. This makes it difficult to scalable apply RL post-training to the self-evolving learning of VLMs and continuously improve model performance.

Existing methods, through synthetic data~\cite{georeasoning} and self-rewarding mechanisms~\cite{li2025self}, can effectively expand data, reduce manual annotation costs, and enhance the visual perception and multi-modal reasoning capabilities of VLMs. Yet, multi-modal synthetic questions tend to converge to limited, repetitive distributions, and self-rewarding mechanisms are mostly used to improve models’ visual perception capabilities, with their training objectives lacking reasonable alignment with reasoning tasks in specialized domains. These methods thus struggle to sustainably enhance the performance of VLMs in specialized domains. A fundamental reason for this failure is that most approaches only utilize question-answering training data and verify the correctness of model outputs through rule-based methods. This supervision paradigm based on "answering questions—receiving rewards" overlooks the rich contextual information in question stems, making it difficult to stimulate models’ general reasoning thinking within the limited distribution of synthetic questions. Consequently, during RL fine-tuning, the model is incentivized to exploit reward-associated shortcuts rather than to genuinely reason, leading to classic \emph{reward hacking}, where models exploit high-reward shortcuts. It results in entropy collapse and diminished exploration, ultimately limiting the capacity of existing methods to deliver generalizable performance in specialized domains. 

Based on the above intuition, we propose \textbf{DoGe} (\textbf{D}ec\textbf{o}uple to \textbf{Ge}neralize), a dual-decoupling framework that restructures the learning process into a two-stage RL training loop. Compared with traditional reinforcement learning post-training methods, as shown in Figure~\ref{fig:intro}, DoGe first trains the model to comprehensively and in-depth understand the contextual information that does not contain the problem itself by decoupling the policy model into two dual components: \textbf{Thinker} and \textbf{Solver}. We have reasonably quantified the reward signal for this process. In the second stage, DoGe requires the model to apply the acquired high-level knowledge to solve the original problem, thereby fostering the model to develop corresponding reasoning capabilities. Aligned with the logic of human cognitive processes in psychology, DoGe establishes a complete evolutionary cycle of "learning-application-internalized understanding," providing insights for realizing advanced agents capable of continuous self-learning. We validate DoGe across seven benchmarks spanning domain-specific reasoning and general visual capabilities. Compared to the baseline method, DoGe demonstrates more stable performance growth and better generalization during the iterative self-evolution process. Analysis of policy entropy confirms that DoGe effectively enhances the model's explorability and avoids potential entropy collapse and reward hacking.

\section{Related Work}
\label{sec:relatedwork}

\subsection{Vision Language Model Post-Training}

While early VLMs primarily relied on supervised pre-training, recent advances demonstrate that post-training strategies can significantly boost generalization. LLaVA~\cite{liu2023llava} pioneers multimodal instruction tuning by generating synthetic image–instruction data with GPT-4. InstructBLIP~\cite{dai2023instructblip} employs diverse multimodal instruction sets and an instruction-aware query transformer. Subsequent works explore modular adaptation, including Mixture-of-Modality Adapters in LaVIN~\cite{luo2023lavin} and dynamic routing through MixLoRA~\cite{shen2024mixlora}. RL fine-tuning has been shown to unlock complex reasoning in LVLMs~\cite{zhai2024rlvlm,zhang2025r1}, improve tasks like captioning, VQA and multimodal dialogue~\cite{xing2025caprl,shen2025vlm}. For example, Huang et al. introduce Vision-R1~\cite{huang2025vision}, adapting DeepSeek-R1-style RL~\cite{guo2025deepseek} to multimodal reasoning: they create a synthetic CoT dataset for a cold-start, then fine-tune with Group Relative Policy Optimization on math problems. OpenVLThinker~\cite{deng2025openvlthinker} alternate supervised fine-tuning (SFT) and RL to surface latent reasoning capabilities in a vision-language model. Other approaches focus on reward design. Vision-SR1~\cite{li2025self}, a self-rewarding RL framework that strengthens visual alignment and reduces hallucinations without external labels. These trends illustrate a growing emphasis on RLVR for VLMs: by leveraging objective correctness signals, these methods improve model alignment (e.g. reducing visual hallucination) and verifiability without needing costly human annotations~\cite{byun2024ares}.

\subsection{Self-Evolving Vision Language Model}

Recent research has focused on enabling self-evolution in vision-language models through reinforcement learning, marking a transition from the "Era of Human Data" to the "Era of Experience". This paradigm shift encompasses three key directions of exploration.
Parameter-update-based methods empower models to autonomously master new skills via environmental interaction: SEAgent~\cite{sun2025seagent} leverages a world state model and curriculum generator to enable computer-use agents to self-evolve in novel software environments without human annotations; VL-Rethinker~\cite{wang2025vl} incentivizes self-reflection by appending triggers like "Let's think again" to model outputs; for unified learning, UniRL~\cite{mao2025unirl} and CoRL~\cite{tarasov2023corl} jointly optimize understanding and generation tasks to achieve emergent cross-modal generalization.
Training-free self-evolution methods enhance capabilities through in-context learning without parameter updates: Training-Free GRPO shifts learning from parameter space to context space by distilling experiential knowledge into external textual libraries injected during inference, achieving fine-tuning-like effects at zero training cost while avoiding overfitting.
Domain-specific frameworks further demonstrate the paradigm’s broad application potential: Q-Insight~\cite{li2025q} applies GRPO to image quality assessment using IoU-based rewards for more accurate, well-reasoned quality scores; Med-R1~\cite{lai2025med} encourages diverse reasoning pathways in medical imaging analysis, achieving cross-modal generalization without annotated rationales. These applications validate the versatility of self-evolution paradigms across creative tasks and safety-critical systems alike.

\begin{figure}[ht]
\centering
\includegraphics[width=1\linewidth]{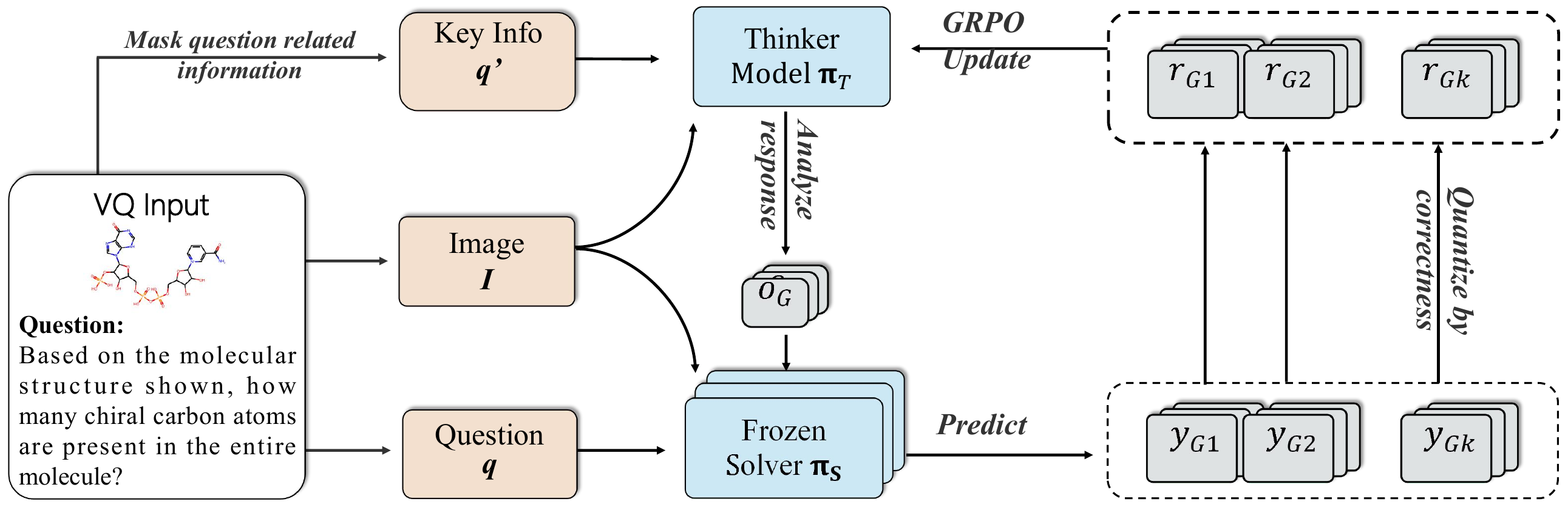} 
\caption{DoGe masks the question part and retains only the context. As shown in the example, DoGe preserves only the molecular image input. The Thinker attempts to conduct in-depth thinking without specific question input, embeds its output into the Solver, and uses the pass rate of the Solver in solving the original question as the quantitative reward criterion for the Thinker.}
\vspace{-2mm}
\label{fig:doge}
\end{figure}

\section{Method}
\label{sec:method}

From a cognitive perspective, the \textbf{DoGe} (\textbf{D}ec\textbf{o}uple to \textbf{Ge}neralize) framework decouples the self-evolving learning process of VLMs into two modules, "Learning" and "Application". It is built on multimodal Group Relative Policy Optimization (GRPO). In this section, we first formalize the task objectives, then introduce our method, followed by a concise multimodal data synthesis pipeline.

\subsection{Preliminaries}
\label{subsec:preliminaries}
We consider a vision-language reasoning task where each training sample is a triplet $(x, q, y)$, consisting of a multimodal context $x$, a question $q$, and a ground-truth answer $y$. A VLM learns a policy $\pi_{\theta}$ that generates a correct answer $y$. In a RL setting, the policy is optimized to maximize the expected reward over a data distribution $\mathcal{D}$:
\begin{equation}
    J(\theta) = \mathbb{E}_{(x,q,y) \sim \mathcal{D}, \hat{y} \sim \pi_\theta(\cdot|x,q)} \left[ r(x,q,\hat{y}) \right],
\end{equation}
where the reward function $r(\cdot)$ typically measures the quality of the generated answer, and is typically grounded in correctness signals derived from the reference answer~$y$.

Traditional reinforcement learning methods are quite sensitive to the diversity and quality of the problem distribution $\mathcal{D}$, making them difficult to scale effectively in multimodal professional domains that lack high-quality data and rely primarily on synthetic data. Inspired by human cognitive processes, we reconsider the problem context information overlooked by reinforcement learning. DoGe decouples the thinking process into two distinct, functional components derived from the same base VLM: a \textbf{Thinker} $\pi_T(a|\tilde{x})$ that performs contextual analysis, and a \textbf{Solver} $\pi_S(\hat{y}|x,q,a)$ that solves the problem based on the Thinker’s analysis. The Thinker receives a question-masked input $\tilde{x} = \text{Mask}(x,q)$, where the question and any direct answer cues are removed, and generates a textual analysis $a$, representing its understanding of the context. The Solver receives the original context $x$, the question $q$, and the Thinker's analysis $a$ to produce the final answer $\hat{y}$. At the beginning of each training round $t$, both policies are initialized from the parameters of the base model from the previous round, $\pi^{(t)}$:
\begin{equation}
\pi_T^{(t)} = \pi_S^{(t)} \leftarrow \pi^{(t)}.
\label{eq:pg_objective}
\end{equation}
This setup forms the foundation of our DoGe approach.

\begin{figure}[ht]
\centering
\includegraphics[width=1\linewidth]{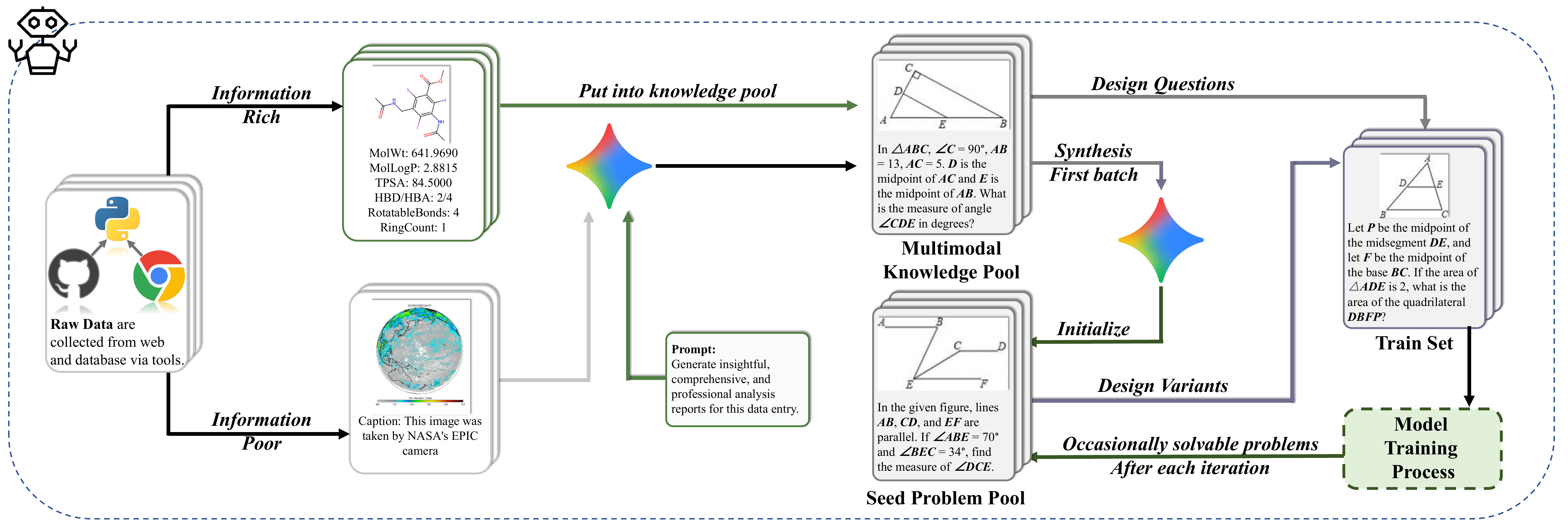} 
\caption{DoGe's data synthesis framework is analogous to the learning process of humans—learning knowledge from the world and then applying it to solve problems. DoGe first collects a large amount of unlabeled data from the web and databases via tools. The data is aggregated into a Multimodal Knowledge Pool. The LVLM transforms it into learnable vision-question-answer pairs. The training data for DoGe consists of the those designed questions and variant problems synthesized from the iteratively updated Seed Problem Pool.}
\label{fig:synthesis}
\end{figure}

\subsection{Decoupling Cognitive Process: 2-Stage RL}

DoGe employs a two-stage Reinforcement Learning framework including \textit{Learning from Context} and \textit{Learning from Application}. The training is guided by the general policy objective defined in Equation~\ref{eq:pg_objective}, whose gradient is formally expressed as:
\begin{equation}
    \nabla_\theta J(\theta) = \mathbb{E}_{\hat{y} \sim \pi_\theta(\cdot|x,q)} \left[ \nabla_\theta \log \pi_\theta(\hat{y}|x,q) r(x,q,\hat{y}) \right].
    \label{eq:pg_gradient}
\end{equation}

\textbf{Learning from Context} In each training round, we first train the model with Learning from Context. As shown in Section~\ref{subsec:preliminaries}, during this phase, we decouple the initial policy network into a learnable \textbf{Thinker} and a parameter-frozen \textbf{Solver}. Rather than directly solving the task, the Thinker is prompted to freely explore the task's contextual information and generate a comprehensive analysis. The Solver addresses the original task by referencing this result, with the solution accuracy serving as its reasonable quantitative reward. Formally, for a given training triplet $(x, q, y)$, we construct a \emph{question-masked} input $\tilde{x} = \mathrm{Mask}(x, q)$, which preserves the essential scenario information while removing explicit question or answer clues. The Thinker then produces a set of candidate contextual analyses:
\begin{equation}
a_k \sim \pi_T^{(t)}(\cdot | \tilde{x}),
\end{equation}
where each $a_k$ describes the Thinker’s interpretation of the problem setup. A frozen copy of the same model, denoted as the Solver $\pi_S^{(t)}$, assess the practical usefulness of these analyses by attempting to solve the original problem conditioned on $a_k$. The rule-base reward function for this stage, $r_{\text{context}}$, is computed with formatting bonus as:
\begin{equation}
r_{\text{context}} = \mathbb{E}_{\hat{y} \sim \pi_S^{(t)}(\cdot | x, q, a_k)} \left[ \mathbf{1}\left[ \hat{y} = y \right] + \beta \cdot r_{\text{format}}(\hat{y}) \right],
\end{equation}
where $r_{\text{format}}(\hat{y})$ is the format reward, which gives a binary reward (0 or 1) by checking if the model output conforms to the "$\langle \text{think} \rangle$  Thinking Process $\langle / \text{think} \rangle$ Final Answer format".
By separating task context from problem-solving, Thinker is encouraged to deeply understand the knowledge domain involved in the task and explore diverse reasoning paths. Solver constructs a self-supervised feedback loop, which measures the value and depth of Thinker's responses. After this stage, the updated Thinker parameters $\pi_T^{(t+1)}$ capture a richer representation of contextual reasoning patterns.

\textbf{Learning from Application} We prompt current Thinker $\pi_T^{(t+1)}$ to apply the learned high-level thinking to solve the original problem, internalizing contextual knowledge into reasoning capabilities. During this training phase, the Thinker is optimized on the original training problems. The reward, $r_{\text{app}}$, is assigned directly based on whether the final generated answer $\hat{y}$ is correct, plus a format bonus:
\begin{equation}
r_{\text{app}} = \mathbb{E}_{\hat{y} \sim \pi_T^{(t+1)} (\cdot | x, q)}\left[\mathbf{1}\left[ \hat{y} = y \right] + \beta \cdot r_{\text{format}}(\hat{y})\right].
\end{equation}
The updated parameters $\pi_T^{(t+1)'}$ are then used to initialize the Solver for the next round, enabling the entire system to self-bootstrap: the improved Thinker enhances context understanding, which in turn empowers the next Solver to perform progressively more generalizable reasoning.

The policy optimization in both stages is performed using the GRPO algorithm. A detailed description of our implementation and the objective function is in Appendix~\ref{sec:grpo_appendix}.

\subsection{Iterative Curriculum Data Synthesis}
DoGe introduces a concise and efficient data synthesis pipeline. It is designed to simulate the process where self-evolving VLMs acquire knowledge, devise challenging reasoning questions, and ultimately internalize these into reasoning capabilities. As illustrated in Figure~\ref{fig:synthesis}, DoGe first leverages tools to collect a large volume of unannotated textual or multimodal raw data from the web and databases. 

The collected raw data varies significantly in information density. For samples with poor information (e.g., only an image and its capture date), we prompt frontier multimodal large models to process these samples and generate expert-level analysis. Additionally, for samples lacking image data, we invoke corresponding tools to generate high-quality images for them. Finally, the samples entering the \textit{Multimodal Knowledge Pool} contain rich domain-specific knowledge and contextual information. To internalize these into the model's capabilities through reinforcement learning fine-tuning, we follow most synthetic data methods and call upon state-of-the-art LVLMs to synthesize reasoning Question-Answer pairs based on these samples.

We additionally maintain a seed problem pool to store high-quality seed problems. In each iteration, we sample from this pool to synthesize challenging variant problems. This method has been proven by prior work to effectively enhance the diversity of training data and consistently and stably apply RL algorithms to improve model performance~\cite{liang2025beyond}. The seed problem pool can be initialized using the training samples synthesized from external information in the first round, or by introducing high-quality seed data. After each iteration, we use the current policy model $\pi_{\theta}$ to calculate the pass rate on the training dataset and select problems that are occasionally solvable by the model (e.g., $0.1 \le \text{pass rate} \le 0.3$) to update the seed problem pool.
In summary, DoGe constructs a complete and effective data synthesis workflow: it collects knowledge within specific domains and annotates it into challenging reasoning questions. Directly applying algorithms such as GRPO and PPO to these questions is the current general approach for realizing self-evolving VLMs through reinforcement fine-tuning. From the perspective of cognitive theory, DoGe reconsiders the overlooked contextual information and explores a new training paradigm for self-evolving agents.

\section{Experiment}
\label{sec:experiment}

\subsection{Experiment Settings}


\subsubsection{Data Preparation}
\label{subsubsec:data-preparation}
We first introduce the details of reinforcement learning training data synthesis. As shown in Figure~\ref{fig:synthesis}, in the iterative curriculum learning framework, we first collected a large amount of unlabeled raw multi-modal data using various tools. This data is then applied within a LVLM-based Agent Workflow to synthesize multi-modal reasoning training data. We selected three domains: \textbf{Multimodal Mathematics}, \textbf{Chemistry}, and \textbf{Earth Science}; the methods for collecting their raw data are as follows:

\noindent\textbf{Multimodal Mathematics} We collected approximately 10k images from Geometry3k~\cite{lu2021inter}, geoqa-plus-train (model's geometric problem-solving ability), and ChartQA-test~\cite{masry2022chartqa} (chart perception, calculation, and mathematical reasoning ability). At the same time, we removed the original high-quality questions from the data to simulate the data domain lacking high-quality manual annotations, retaining only the key information describing the images; for instance, for geometry problems, we kept the description of the image provided in the original question; for charts, as the chart itself is fully observable, we only retained the image itself without any additional text descriptions.

\noindent\textbf{Chemistry} We first obtained the SMILES molecular formulas of compounds by randomly sampling CIDs via the PubChem API; then, using the RDKit library, based on the SMILES formula, we synthesized the corresponding molecular structure images, along with compound-related properties including IUPAC name, TPSA, MolWt, etc. We combined these into a complete piece of information for subsequent training data synthesis. In this stage, we collected related information and molecular structure images for approximately 2500 compounds in total.

\noindent\textbf{Earth Science} We used the Earth Polychromatic Imaging Camera (EPIC) API provided by the National Aeronautics and Space Administration (NASA) to obtain Earth satellite images. The raw data collected only contained: four forms of Earth satellite images (natural, enhanced, aerosol, cloud), and the image capture date. We collected approximately 1500 related data entries in total.

We classify the raw multi-modal data collected from tools and the web into two categories, \textit{Information Rich} and \textit{Information Poor}, based on the information content of their accompanying descriptive text. For samples lacking sufficient information, to alleviate the burden on the LLM during the subsequent problem design process, we first invoked the Gemini-2.5-Flash~\cite{comanici2025gemini} model to generate insightful, comprehensive and professional analysis reports to expand the sample information. Those processed samples are integrated into the Multimodal Knowledge Pool, which serves as the foundation for self-evolving VLMs to learn and evolve from domains lacking high-quality manually designed problems.

DoGe's data synthesis framework maintains and updates the \textit{Seed Problem Pool} module throughout multiple iterations of VLM self-evolution. This module stores reasoning problems occasionally solvable by the policy model from the training data, such problems have been proven by previous curriculum learning method to accelerate the model's RL process and improve model performance. Except that approximately 1.2k modified questions from SMolInstruct~\cite{yu2024llasmol} were used to initialize the chemistry domain data, both earth science and multimodal mathematics adopted subsets of problems synthesized based on the \textit{Multimodal Knowledge Pool} in the first round.

In each iteration, by sampling from the \textit{Multimodal Knowledge Pool} and the \textit{Seed Problem Pool}, we prompt large models to synthesize challenging reasoning training problems. Before and after training, the policy model samples responses multiple times, and overly simple training problems are filtered out based on accuracy rates. The remaining problems are used for model training and updating the \textit{Seed Problem Pool}.

\begin{table*}[ht]
    \small
    \centering
    \caption{DoGe vs. Baseline. All experiments adopted the same setup, and the specific details of the experimental parameters can be referred to in the Appendix. For Qwen2.5VL-7B, as stated in Section~\ref{sec:experiment}, we apply GRPO Algorithm with format reward only for 15 steps to better follow instructions.}
    \setlength{\tabcolsep}{0.2mm}{
   \begin{tabular}{l ccc cccc c}
    \toprule
   \multirow{2}{*}{\textbf{Method}} & \multicolumn{3}{c}{\bfseries General Visual Reasoning \& Hall} & \multicolumn{4}{c}{\bfseries Specific Domain} & \\
     \cmidrule(lr){2-4} \cmidrule(lr){5-8}
         & MMMU & MMStar & HallBench & MathVision & MathVista & ChemBench & MSEarthMCQ & \textbf{Avg.} \\
        \midrule
        \rowcolor[HTML]{EFEFEF}
        \multicolumn{9}{c}{\textit{3B-level Models}} \\
        \midrule
        InternVL2.5-2B~\cite{chen2024internvl} & 43.6 & 53.7 & 42.6 & 13.5 & 51.3 & - & - & - \\
        Visionary-3B~\cite{xia2025visionary} & 40.7 & 50.5 & 59.8 & 17.1 & 54.7 & 40.8 & 38.2 & 43.1 \\
        Qwen2.5VL-3B* & 41.0 & 49.3 & 60.6 & 18.7 & 48.8 & 43.4 & 40.8 & 43.2 \\
        \hline
        \textit{Ours Method} \\
        DoGe-3B (Iter1) & 46.6 & 54.5 & 61.5 & 21.7 & \textbf{57.9} & 45.8 & \textbf{48.3} & 48.0 \\
        DoGe-3B (Iter2) & 48.9 & 52.5 & \textbf{62.5} & 23.1 & 54.2 & \textbf{47.7} & 46.2 & 47.9 \\
        DoGe-3B (Iter3) & \textbf{50.2} & \textbf{54.7} & 61.8 & \textbf{24.2} & 57.0 & 46.9 & 47.3 & \textbf{48.9} \\
        \midrule
        $\Delta$ max(vs. Base Model) & $+9.2$ & $+5.4$ & $+1.9$ & $+5.5$ & $+9.1$ & $+4.3$ & $+7.5$ & $+5.7$ \\
        \midrule
        \rowcolor[HTML]{EFEFEF}
        \multicolumn{9}{c}{\textit{7B-level Models}} \\
        InternVL2.5-8B & 48.9 & 62.8 & 50.1 & 22.0 & 64.4 & - & - & - \\
        Vision-R1-7B~\cite{huang2025vision} & 46.9 & 60.8 & 66.7 & \textbf{29.0} & 68.5 & 46.0 & 44.1 & 51.7\\
        Qwen2.5VL-7B* & 49.9 & 60.7 & 66.3 & 23.6 & 64.1 & 48.6 & 43.3 & 50.9 \\  
        \hline
        \textit{Ours Method} \\
        DoGe-7B (Iter1) & 53.1 & \textbf{63.2} & 54.4 & 24.3 & 62.1 & 48.7 & 46.4 & 50.3 \\
        DoGe-7B (Iter2) & 50.9 & 60.0 & \textbf{68.3} & 25.3 & \textbf{68.8} & \textbf{49.0} & \textbf{46.5} & 52.7 \\
        DoGe-7B (Iter3) & \textbf{53.6} & 63.0 & 68.0 & 25.2 & 68.3 & 48.5 & 45.8 & \textbf{53.2} \\
        \midrule
        $\Delta$ max(vs. Base Model) & $+3.7$ & $+2.5$ & $+2.0$ & $+1.7$ & $+4.7$ & $+0.4$ & $+3.2$ & $+2.3$ \\
        \bottomrule
    \end{tabular}}
    \label{tab:performance1}
\end{table*}

\subsubsection{Benchmarks}
We a comprehensive suite of seven benchmark tests to evaluate the capabilities of VLMs in two aspects: 1) reasoning ability in specific scarce data domains; 2) general visual reasoning ability and hallucination.

\noindent\textbf{Reasoning Ability in Specific Scarce Data Domains:}
\textbf{(1) MSEarthMCQ}~\cite{bai2025intern} consists of 2784 samples, encompassing the five major spheres of Earth science: atmosphere, cryosphere, hydrosphere, lithosphere, and biosphere. It measures the ability of VLMs to solve graduate-level Earth science reasoning tasks. \textbf{(2) ChemBench}, as a text-only non-multimodal dataset, covers nine chemical tasks based on text: molecular generation, name conversion, property prediction, temperature prediction, molecular description, yield prediction, solvent prediction, retrosynthetic analysis, and product prediction. Previous studies have shown that reinforcement learning training for VLMs, while improving their image reasoning ability, can impair the model's text reasoning ability to a certain extent~\cite{zhang2024wings,ratzlaff2025training}; we aim to illustrate through the comparison between our method and baseline on this dataset that our method can alleviate this problem to a certain degree. \textbf{(3) MathVista}~\cite{lu2023mathvista} and \textbf{(4) MathVision}~\cite{wang2024measuring}; we selected the test and testmini subsets of these two benchmarks respectively. These two benchmarks are not limited to multimodal mathematical problems, including natural images, geometric diagrams, abstract scenes, etc., and systematically evaluate the model's mathematical reasoning ability in visual contexts. 

\noindent\textbf{General Visual Reasoning Ability and Hallucination:} 
\textbf{(5) MMMU}~\cite{yue2024mmmu} evaluates the ability of multimodal models on a large number of interdisciplinary tasks that require university-level subject knowledge and deliberate reasoning. It covers six core domains: Art \& Design, Business, Science, Health \& Medicine, Humanities \& Social Sciences, and Technology \& Engineering. \textbf{(6) MMStar}~\cite{chen2024we} contains 1500 carefully selected visually critical samples. It addresses the problems of visual redundancy and data leakage in existing evaluations, improving the accuracy of multimodal performance assessment. \textbf{(7) HallusionBench}~\cite{guan2024hallusionbench}. Previous studies have shown that VLMs have a serious hallucination problem, that is, they over-rely on language priors and abandon thinking based on images. This dataset contains 951 image-text data pairs, which systematically evaluate the visual hallucination phenomenon.

\subsubsection{Training Details}
We select Qwen2.5VL-7B-Instruct and Qwen2.5VL-3B-Instruct~\cite{bai2025qwen2} as our base models, and conduct training on models of different scales to demonstrate the general adaptability of our method. We use 8$\times$A100 GPUs, and both DoGe's training phases are based on the GRPO algorithm, with the training code modified from the verl framework~\cite{sheng2024hybridflow}. The training steps for Thinker $\pi_T$ are 100, and the training steps when applying GRPO annealing in the second stage are 150. For the base model Qwen2.5VL-7B-Instruct, both stages are set to 150 steps. We set the train batch sizes to 64 and 48 respectively, with a maximum response length of 4096. For DoGe training phase 2, we sample 8 responses per problem, while for DoGe training phase 1, we sample 4 responses; for each response, we output 4 answers through Solver $\pi_S$ to estimate its reward. Following the work of DAPO, we decoupled the clip $\epsilon$ parameter and adopted different values for training phase 1 and 2. For specific details, please refer to the Appendix~\ref{sec:experiment_appendix}.

\begin{table*}[ht]
   \small
    \centering
    \vspace{-2mm}
    \caption{Ablation between DoGe and naive GRPO.}
    \setlength{\tabcolsep}{1.2mm}{
   \begin{tabular}{l cc cccc c}
    \toprule
   \multirow{2}{*}{\textbf{Method}} & \multicolumn{2}{c}{\bfseries General Visual Reasoning} & \multicolumn{4}{c}{\bfseries Specific Domain} & \\
     \cmidrule(lr){2-3} \cmidrule(lr){4-7}
         & MMMU & MMStar & MathVision & MathVista & ChemBench & MSEarthMCQ & \textbf{Avg.} \\
        \midrule
        Our Method (Iter1) &  53.1 &  63.2 &  24.3 & 62.1 &  48.7 &  46.4 & 49.6 \\
        \quad $\vdash$ w/o DoGe & 52.9 & 62.9 & 23.2 & 66.1 & 48.4 & 46.0 & 49.9\\
        \hline
        Our Method (Iter2) &  50.9 &  60.0 &  25.3 &  68.8 &  49.0 &  46.5 &  50.1 \\
        \quad $\vdash$ w/o DoGe & 48.2 & 57.0 & 18.0 & 62.4 & 47.6 & 46.1 & 46.6 \\
        \hline
        Our Method (Iter3) &  53.6 &  63.0 &  25.2 &  68.3 &  48.5 & 45.8 &  50.7 \\
        \quad $\vdash$ w/o DoGe & 51.9 & 62.8 & 24.8 & 65.2 & 48.4 & 48.1 & 50.2 \\
        \bottomrule
    \end{tabular}}
    \vspace{-2mm}
    \label{tab:ablation1}
\end{table*}

\subsection{Main Results}
We used vllm~\cite{kwon2023efficient} to test the model's capabilities, with the maximum response length set to 4096, and adopted a rule-based reward to evaluate the model's accuracy using mean@4. To better enable the base model to follow instructions, we performed 15 steps of RL training on both Qwen2.5VL-7B-Instruct and Qwen2.5VL-3B-Instruct. Specifically, we only provided format rewards, allowing the model to output answers in the correct format without compromising its capabilities.

We conducted a total of 3 iterations. As shown in the table~\ref{tab:performance1}, our method enables VLM to stably achieve self-evolution and performance improvement, enhancing the model's performance across all benchmark tests. The 3B and 7B series models achieved an average performance increase of 5.7\% and 2.3\% respectively across 7 benchmark tests. By removing the question and forcing the model to analyze multimodal context during training phase 1, DoGe avoid the flaw of traditional multimodal reasoning models relying on text priors to answer questions~\cite{liu2025thinkingseeingassessingamplified,li2025self}, strengthened the model's visual perception ability, and achieved an average improvement of 2.0\% on HallBench. While reducing model hallucination, DoGe improves the model's reasoning ability in specific domains and effectively generalizes to general domains. Notably, the prior knowledge distribution(Multimodal Knowledge Pool) we used to construct synthetic problems is quite limited, but DoGe still effectively enhanced model performance by decoupling the cognitive process into a "learning-application" cycle. We compared two previous works, visionary-3b and vision-r1, and our method still demonstrates superior performance, leading on multiple benchmark tests.

\subsection{Ablation study on Our method}

In this section, starting from the experimental results and comparing with the baseline reinforcement learning post-training, we will explain how DoGe enhances model generalization through decoupling cognitive processes and leveraging two-stage RL training. We first analyze the model's policy entropy and compare training output logs; then we analyze the model's performance after removing the 2-stage DoGe training strategy.


\begin{figure}[ht]
\centering
\begin{minipage}{0.5\linewidth}
    \centering
    \includegraphics[width=\linewidth]{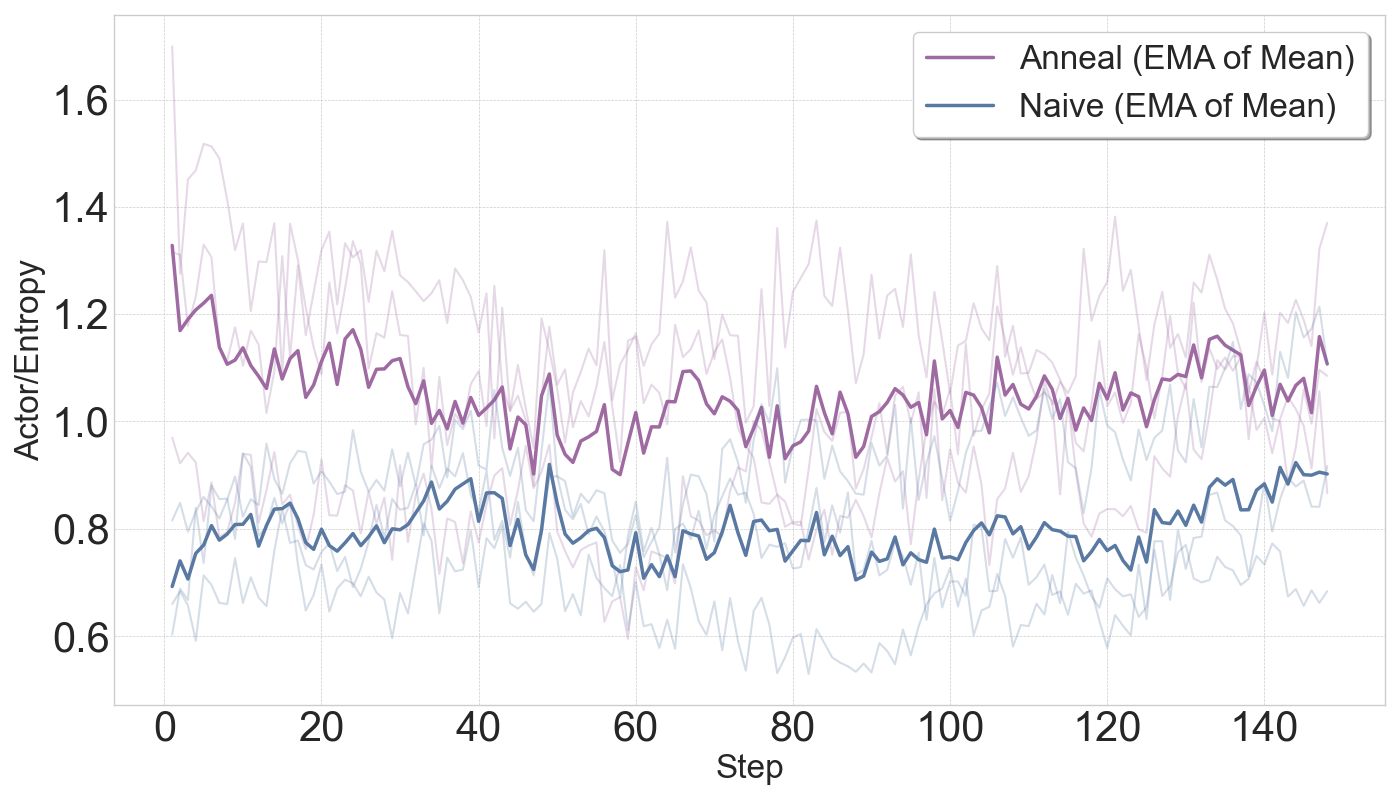}
    \captionof{figure}{Average Entropy (DoGe vs. Baseline) during training. "Anneal" refers to DoGe's RL stage 2. Compared to the baseline, DoGe exhibits a higher initial policy entropy during training and consistently maintains greater exploration.}
    \label{fig:entropy}
\end{minipage}
\hfill
\begin{minipage}{0.46\linewidth}
    \centering
    \includegraphics[width=\linewidth]{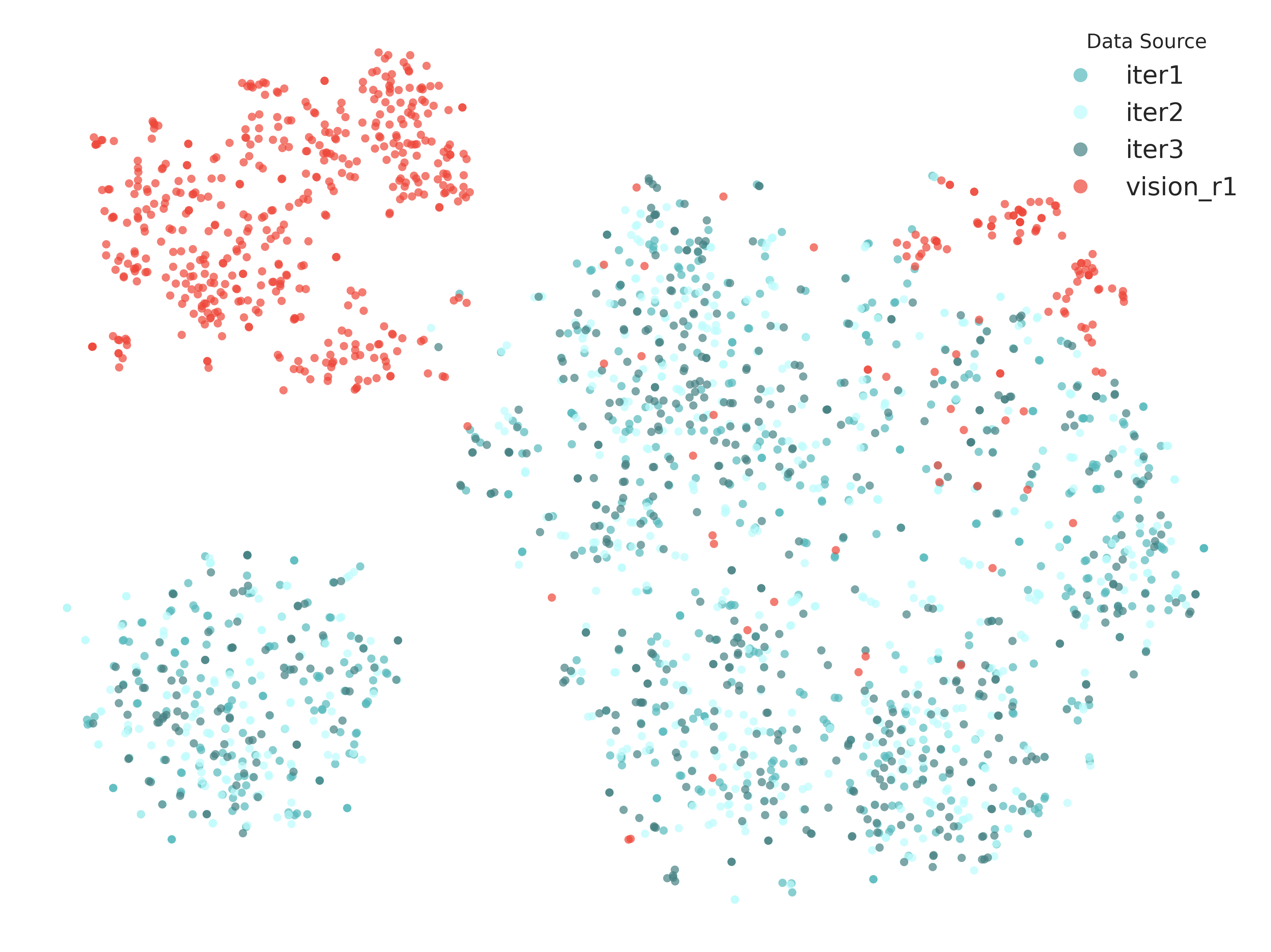}
    \captionof{figure}{Distribution comparison on the training data in the mathematical field. We select training data's subset of Vision-R1 and ours method. The visualization results presented by applying text-embedding-3-large embedding to the text part of the problem and then performing t-SNE dimensionality reduction.}
    \label{fig:data1}
\end{minipage}
\end{figure}

\textbf{Analysis on Policy Entropy} We found that training Thinker$(\pi_T)$ in advance in the training phase 1 before using GRPO annealing in the training phase 2 can effectively increase the initial policy entropy of the model and maintain a higher entropy value during the training process. As shown in Figure~\ref{fig:entropy}, we analyzed the policy entropy during training for directly applying naive GRPO algorithm and for applying DoGe's decoupled 2-stage RL. We averaged the curves of 3 iterations and applied EMA smoothing. It can be seen from the figure that our method can effectively increase the initial policy entropy of subsequent reinforcement learning training and maintain high exploration during training. The first training phase motivates the model to develop a deep understanding of contextual information, enabling it to pre-learn high level thinking patterns rather than merely repeating actions with high reward. This effectively avoids the problem where the baseline method, due to excessively low entropy and reduced exploration ability, struggles to effectively learn more generalizable universal reasoning paradigms.

\textbf{Analysis on Performance} Furthermore, by analyzing the performance changes during iterations, as shown in Table~\ref{tab:ablation1}, DoGe stably outperforms the baseline method across various benchmarks, especially in reasoning-intensive tasks such as multimodal mathematics and general reasoning. More importantly, our method stabilizes multi-round self-evolving learning. Figure~\ref{fig:intro} shows the performance changes of our method and the baseline algorithm during multiple iterations. Our method can stably improve model performance, while the baseline algorithm exhibits significant fluctuations.
Intuitively, the standard GRPO algorithm is highly sensitive to data quality. In iteration 2, the model may engage in Reward Hacking due to low-quality data, while simultaneously reducing policy entropy, leading to impaired model capabilities. However, because our method harmlessly expands the model's policy entropy, it can robustly learn more robust and generalizable reasoning strategies even with fluctuations in the quality of training data, thereby alleviating this issue.

\subsection{Analysis on Data Synthesis}
In this section, we compare the distribution of synthetic training data with high-quality manually annotated multimodal training data.
We selected partial training data from 3 iterations and randomly chose partial training data used by Vision-R1. We converted the text part of this data (i.e., question design) into embedding vectors using OpenAI text-embedding-3-large to obtain visualization results. As shown in Figure~\ref{fig:data1}, the first figure on the top left displays the distribution comparison between the math part of the synthetic training data and the manually annotated data. It can be seen that although the synthetic data and the manually annotated data share similar question design contexts (i.e., geometric images, charts, and corresponding descriptions), the data synthesized by the model has a wider range of question surface designs. This means that manually annotated data has higher requirements for the diversity and quality of images; otherwise, similar image-text questions will cause the model to continuously output similar actions during the reinforcement learning process. These actions are continuously rewarded, leading to a decrease in policy entropy and further affecting generalization. Further more, this indicates that there is more unused information in the question context, and a single question is insufficient to measure the model's understanding of the context.


\section{Conclusion}
\label{sec:conclusion}
In this paper, We propose the DoGe framework, a two-stage reinforcement learning framework that reasonably decouples cognitive processes. It prompts the model to first understand knowledge and learn top-level thinking, then apply knowledge and internalize reasoning capabilities, thereby enabling the self-evolving of VLMs in the era of experience. We first prompt the Thinker to comprehensively analyze the problem contextual information. Then, we embed non-quantitative outputs into Solver with frozen parameters in the form of prompts to solve the original problems, thereby obtaining quantitative rewards to update the model.
We demonstrate through experiments that DoGe can effectively improve the generalization ability of the model, enhance the explorability of reinforcement learning phase, reduce the model's sensitivity to data quality, and stabilize training.
We argue that DoGe provides a reliable approach to realizing self-evolving VLMs and facilitating the continuous improvement of models. Especially for special domain tasks lacking high-quality data, this method can effectively mitigate the negative impacts caused by the quality of the data itself, enabling the model to learn robust and generalizable reasoning patterns.


\clearpage
\newpage
\bibliographystyle{plainnat}
\setcitestyle{numbers}
\bibliography{ref}

\begin{thebibliography}{46}
\providecommand{\natexlab}[1]{#1}
\providecommand{\url}[1]{\texttt{#1}}
\expandafter\ifx\csname urlstyle\endcsname\relax
  \providecommand{\doi}[1]{doi: #1}\else
  \providecommand{\doi}{doi: \begingroup \urlstyle{rm}\Url}\fi

\bibitem[Bai et~al.(2025{\natexlab{a}})Bai, Cai, Cao, Cao, Cao, Chen, Chen, Chen, Chen, Chen, et~al.]{bai2025intern}
Lei Bai, Zhongrui Cai, Yuhang Cao, Maosong Cao, Weihan Cao, Chiyu Chen, Haojiong Chen, Kai Chen, Pengcheng Chen, Ying Chen, et~al.
\newblock Intern-s1: A scientific multimodal foundation model.
\newblock \emph{arXiv preprint arXiv:2508.15763}, 2025{\natexlab{a}}.

\bibitem[Bai et~al.(2025{\natexlab{b}})Bai, Chen, Liu, Wang, Ge, Song, Dang, Wang, Wang, Tang, et~al.]{bai2025qwen2}
Shuai Bai, Keqin Chen, Xuejing Liu, Jialin Wang, Wenbin Ge, Sibo Song, Kai Dang, Peng Wang, Shijie Wang, Jun Tang, et~al.
\newblock Qwen2. 5-vl technical report.
\newblock \emph{arXiv preprint arXiv:2502.13923}, 2025{\natexlab{b}}.

\bibitem[Byun et~al.(2024)Byun, Chun, Kil, and Perrault]{byun2024ares}
Ju-Seung Byun, Jiyun Chun, Jihyung Kil, and Andrew Perrault.
\newblock Ares: Alternating reinforcement learning and supervised fine-tuning for enhanced multi-modal chain-of-thought reasoning through diverse ai feedback.
\newblock \emph{arXiv preprint arXiv:2407.00087}, 2024.

\bibitem[Chen et~al.(2024{\natexlab{a}})Chen, Li, Dong, Zhang, Zang, Chen, Duan, Wang, Qiao, Lin, et~al.]{chen2024we}
Lin Chen, Jinsong Li, Xiaoyi Dong, Pan Zhang, Yuhang Zang, Zehui Chen, Haodong Duan, Jiaqi Wang, Yu~Qiao, Dahua Lin, et~al.
\newblock Are we on the right way for evaluating large vision-language models?
\newblock \emph{Advances in Neural Information Processing Systems}, 37:\penalty0 27056--27087, 2024{\natexlab{a}}.

\bibitem[Chen et~al.(2024{\natexlab{b}})Chen, Wu, Wang, Su, Chen, Xing, Zhong, Zhang, Zhu, Lu, et~al.]{chen2024internvl}
Zhe Chen, Jiannan Wu, Wenhai Wang, Weijie Su, Guo Chen, Sen Xing, Muyan Zhong, Qinglong Zhang, Xizhou Zhu, Lewei Lu, et~al.
\newblock Internvl: Scaling up vision foundation models and aligning for generic visual-linguistic tasks.
\newblock In \emph{Proceedings of the IEEE/CVF conference on computer vision and pattern recognition}, pages 24185--24198, 2024{\natexlab{b}}.

\bibitem[Comanici et~al.(2025)Comanici, Bieber, Schaekermann, Pasupat, Sachdeva, Dhillon, Blistein, Ram, Zhang, Rosen, et~al.]{comanici2025gemini}
Gheorghe Comanici, Eric Bieber, Mike Schaekermann, Ice Pasupat, Noveen Sachdeva, Inderjit Dhillon, Marcel Blistein, Ori Ram, Dan Zhang, Evan Rosen, et~al.
\newblock Gemini 2.5: Pushing the frontier with advanced reasoning, multimodality, long context, and next generation agentic capabilities.
\newblock \emph{arXiv preprint arXiv:2507.06261}, 2025.

\bibitem[Dai et~al.(2023)Dai, Li, Li, Tiong, Zhao, Wang, Li, Fung, and Hoi]{dai2023instructblip}
Wenliang Dai, Junnan Li, Dongxu Li, Anthony Tiong, Junqi Zhao, Weisheng Wang, Boyang Li, Pascale~N Fung, and Steven Hoi.
\newblock Instructblip: Towards general-purpose vision-language models with instruction tuning.
\newblock \emph{Advances in neural information processing systems}, 36:\penalty0 49250--49267, 2023.

\bibitem[Deng et~al.(2025)Deng, Bansal, Yin, Peng, Wang, and Chang]{deng2025openvlthinker}
Yihe Deng, Hritik Bansal, Fan Yin, Nanyun Peng, Wei Wang, and Kai-Wei Chang.
\newblock Openvlthinker: An early exploration to complex vision-language reasoning via iterative self-improvement.
\newblock \emph{arXiv preprint arXiv:2503.17352}, 2025.

\bibitem[Guan et~al.(2024)Guan, Liu, Wu, Xian, Li, Liu, Wang, Chen, Huang, Yacoob, et~al.]{guan2024hallusionbench}
Tianrui Guan, Fuxiao Liu, Xiyang Wu, Ruiqi Xian, Zongxia Li, Xiaoyu Liu, Xijun Wang, Lichang Chen, Furong Huang, Yaser Yacoob, et~al.
\newblock Hallusionbench: an advanced diagnostic suite for entangled language hallucination and visual illusion in large vision-language models.
\newblock In \emph{Proceedings of the IEEE/CVF Conference on Computer Vision and Pattern Recognition}, pages 14375--14385, 2024.

\bibitem[Guo et~al.(2025{\natexlab{a}})Guo, Yang, Zhang, Song, Zhang, Xu, Zhu, Ma, Wang, Bi, et~al.]{guo2025deepseek}
Daya Guo, Dejian Yang, Haowei Zhang, Junxiao Song, Ruoyu Zhang, Runxin Xu, Qihao Zhu, Shirong Ma, Peiyi Wang, Xiao Bi, et~al.
\newblock Deepseek-r1: Incentivizing reasoning capability in llms via reinforcement learning.
\newblock \emph{arXiv preprint arXiv:2501.12948}, 2025{\natexlab{a}}.

\bibitem[Guo et~al.(2025{\natexlab{b}})Guo, Zhang, Chen, Ji, Wang, Hu, and Chen]{zhai2024rlvlm}
Yanjiang Guo, Jianke Zhang, Xiaoyu Chen, Xiang Ji, Yen-Jen Wang, Yucheng Hu, and Jianyu Chen.
\newblock Improving vision-language-action model with online reinforcement learning.
\newblock \emph{arXiv preprint arXiv:2501.16664}, 2025{\natexlab{b}}.

\bibitem[Huang et~al.(2025)Huang, Jia, Zhai, Cao, Ye, Zhao, Xu, Hu, and Lin]{huang2025vision}
Wenxuan Huang, Bohan Jia, Zijie Zhai, Shaosheng Cao, Zheyu Ye, Fei Zhao, Zhe Xu, Yao Hu, and Shaohui Lin.
\newblock Vision-r1: Incentivizing reasoning capability in multimodal large language models.
\newblock \emph{arXiv preprint arXiv:2503.06749}, 2025.

\bibitem[Kwon et~al.(2023)Kwon, Li, Zhuang, Sheng, Zheng, Yu, Gonzalez, Zhang, and Stoica]{kwon2023efficient}
Woosuk Kwon, Zhuohan Li, Siyuan Zhuang, Ying Sheng, Lianmin Zheng, Cody~Hao Yu, Joseph~E. Gonzalez, Hao Zhang, and Ion Stoica.
\newblock Efficient memory management for large language model serving with pagedattention.
\newblock In \emph{Proceedings of the ACM SIGOPS 29th Symposium on Operating Systems Principles}, 2023.

\bibitem[Lai et~al.(2025)Lai, Zhong, Li, Zhao, Li, Psounis, and Yang]{lai2025med}
Yuxiang Lai, Jike Zhong, Ming Li, Shitian Zhao, Yuheng Li, Konstantinos Psounis, and Xiaofeng Yang.
\newblock Med-r1: Reinforcement learning for generalizable medical reasoning in vision-language models.
\newblock \emph{arXiv preprint arXiv:2503.13939}, 2025.

\bibitem[Li et~al.(2024)Li, Ma, Wang, Ye, Cheng, Tang, Zhang, Duan, Zuo, Yang, et~al.]{shen2024mixlora}
Dengchun Li, Yingzi Ma, Naizheng Wang, Zhengmao Ye, Zhiyuan Cheng, Yinghao Tang, Yan Zhang, Lei Duan, Jie Zuo, Cal Yang, et~al.
\newblock Mixlora: Enhancing large language models fine-tuning with lora-based mixture of experts.
\newblock \emph{arXiv preprint arXiv:2404.15159}, 2024.

\bibitem[Li et~al.(2025{\natexlab{a}})Li, Zhang, Zhao, Zhang, Li, Zhang, and Zhang]{li2025q}
Weiqi Li, Xuanyu Zhang, Shijie Zhao, Yabin Zhang, Junlin Li, Li~Zhang, and Jian Zhang.
\newblock Q-insight: Understanding image quality via visual reinforcement learning.
\newblock \emph{arXiv preprint arXiv:2503.22679}, 2025{\natexlab{a}}.

\bibitem[Li et~al.(2025{\natexlab{b}})Li, Yu, Huang, Liu, Liang, Liu, Che, Yu, Boyd-Graber, Mi, et~al.]{li2025self}
Zongxia Li, Wenhao Yu, Chengsong Huang, Rui Liu, Zhenwen Liang, Fuxiao Liu, Jingxi Che, Dian Yu, Jordan Boyd-Graber, Haitao Mi, et~al.
\newblock Self-rewarding vision-language model via reasoning decomposition.
\newblock \emph{arXiv preprint arXiv:2508.19652}, 2025{\natexlab{b}}.

\bibitem[Liang et~al.(2025)Liang, Li, Gong, Shen, Wu, Guo, and Chen]{liang2025beyond}
Xiao Liang, Zhongzhi Li, Yeyun Gong, Yelong Shen, Ying~Nian Wu, Zhijiang Guo, and Weizhu Chen.
\newblock Beyond pass@ 1: Self-play with variational problem synthesis sustains rlvr.
\newblock \emph{arXiv preprint arXiv:2508.14029}, 2025.

\bibitem[Liu et~al.(2025)Liu, Xu, Wei, Wu, Zou, Wang, Zhou, and Liu]{liu2025thinkingseeingassessingamplified}
Chengzhi Liu, Zhongxing Xu, Qingyue Wei, Juncheng Wu, James Zou, Xin~Eric Wang, Yuyin Zhou, and Sheng Liu.
\newblock More thinking, less seeing? assessing amplified hallucination in multimodal reasoning models, 2025.
\newblock URL \url{https://arxiv.org/abs/2505.21523}.

\bibitem[Liu et~al.(2023)Liu, Li, Wu, and Lee]{liu2023llava}
Haotian Liu, Chunyuan Li, Qingyang Wu, and Yong~Jae Lee.
\newblock Visual instruction tuning.
\newblock \emph{Advances in neural information processing systems}, 36:\penalty0 34892--34916, 2023.

\bibitem[Lu et~al.(2021)Lu, Gong, Jiang, Qiu, Huang, Liang, and Zhu]{lu2021inter}
Pan Lu, Ran Gong, Shibiao Jiang, Liang Qiu, Siyuan Huang, Xiaodan Liang, and Song-Chun Zhu.
\newblock Inter-gps: Interpretable geometry problem solving with formal language and symbolic reasoning.
\newblock \emph{arXiv preprint arXiv:2105.04165}, 2021.

\bibitem[Lu et~al.(2023)Lu, Bansal, Xia, Liu, Li, Hajishirzi, Cheng, Chang, Galley, and Gao]{lu2023mathvista}
Pan Lu, Hritik Bansal, Tony Xia, Jiacheng Liu, Chunyuan Li, Hannaneh Hajishirzi, Hao Cheng, Kai-Wei Chang, Michel Galley, and Jianfeng Gao.
\newblock Mathvista: Evaluating mathematical reasoning of foundation models in visual contexts.
\newblock \emph{arXiv preprint arXiv:2310.02255}, 2023.

\bibitem[Luo et~al.(2023)Luo, Zhou, Ren, Chen, Sun, and Ji]{luo2023lavin}
Gen Luo, Yiyi Zhou, Tianhe Ren, Shengxin Chen, Xiaoshuai Sun, and Rongrong Ji.
\newblock Cheap and quick: Efficient vision-language instruction tuning for large language models.
\newblock volume~36, pages 29615--29627, 2023.

\bibitem[Mao et~al.(2025)Mao, Yang, and Shou]{mao2025unirl}
Weijia Mao, Zhenheng Yang, and Mike~Zheng Shou.
\newblock Unirl: Self-improving unified multimodal models via supervised and reinforcement learning.
\newblock \emph{arXiv preprint arXiv:2505.23380}, 2025.

\bibitem[Masry et~al.(2022)Masry, Do, Tan, Joty, and Hoque]{masry2022chartqa}
Ahmed Masry, Xuan~Long Do, Jia~Qing Tan, Shafiq Joty, and Enamul Hoque.
\newblock Chartqa: A benchmark for question answering about charts with visual and logical reasoning.
\newblock In \emph{Findings of the association for computational linguistics: ACL 2022}, pages 2263--2279, 2022.

\bibitem[Parashar et~al.(2024)Parashar, Lin, Liu, Dong, Li, Ramanan, Caverlee, and Kong]{parashar2024neglected}
Shubham Parashar, Zhiqiu Lin, Tian Liu, Xiangjue Dong, Yanan Li, Deva Ramanan, James Caverlee, and Shu Kong.
\newblock The neglected tails in vision-language models.
\newblock In \emph{Proceedings of the IEEE/CVF Conference on Computer Vision and Pattern Recognition}, pages 12988--12997, 2024.

\bibitem[Ratzlaff et~al.(2025)Ratzlaff, Luo, Su, Lal, and Howard]{ratzlaff2025training}
Neale Ratzlaff, Man Luo, Xin Su, Vasudev Lal, and Phillip Howard.
\newblock Training-free mitigation of language reasoning degradation after multimodal instruction tuning.
\newblock In \emph{Proceedings of the AAAI Symposium Series}, volume~5, pages 384--388, 2025.

\bibitem[Schulman et~al.(2017)Schulman, Wolski, Dhariwal, Radford, and Klimov]{schulman2017proximal}
John Schulman, Filip Wolski, Prafulla Dhariwal, Alec Radford, and Oleg Klimov.
\newblock Proximal policy optimization algorithms.
\newblock \emph{arXiv preprint arXiv:1707.06347}, 2017.

\bibitem[Shao et~al.(2024)Shao, Wang, Zhu, Xu, Song, Bi, Zhang, Zhang, Li, Wu, et~al.]{shao2024deepseekmath}
Zhihong Shao, Peiyi Wang, Qihao Zhu, Runxin Xu, Junxiao Song, Xiao Bi, Haowei Zhang, Mingchuan Zhang, YK~Li, Yang Wu, et~al.
\newblock Deepseekmath: Pushing the limits of mathematical reasoning in open language models.
\newblock \emph{arXiv preprint arXiv:2402.03300}, 2024.

\bibitem[Shen et~al.(2025)Shen, Liu, Li, Fang, Ma, Liao, Shen, Zhang, Zhao, Zhang, et~al.]{shen2025vlm}
Haozhan Shen, Peng Liu, Jingcheng Li, Chunxin Fang, Yibo Ma, Jiajia Liao, Qiaoli Shen, Zilun Zhang, Kangjia Zhao, Qianqian Zhang, et~al.
\newblock Vlm-r1: A stable and generalizable r1-style large vision-language model.
\newblock \emph{arXiv preprint arXiv:2504.07615}, 2025.

\bibitem[Sheng et~al.(2024)Sheng, Zhang, Ye, Wu, Zhang, Zhang, Peng, Lin, and Wu]{sheng2024hybridflow}
Guangming Sheng, Chi Zhang, Zilingfeng Ye, Xibin Wu, Wang Zhang, Ru~Zhang, Yanghua Peng, Haibin Lin, and Chuan Wu.
\newblock Hybridflow: A flexible and efficient rlhf framework.
\newblock \emph{arXiv preprint arXiv: 2409.19256}, 2024.

\bibitem[Silver and Sutton(2025)]{silver2025welcome}
David Silver and Richard~S Sutton.
\newblock Welcome to the era of experience.
\newblock \emph{Google AI}, 1, 2025.

\bibitem[Sun et~al.(2025)Sun, Liu, Zang, Cao, Dong, Wu, Lin, and Wang]{sun2025seagent}
Zeyi Sun, Ziyu Liu, Yuhang Zang, Yuhang Cao, Xiaoyi Dong, Tong Wu, Dahua Lin, and Jiaqi Wang.
\newblock Seagent: Self-evolving computer use agent with autonomous learning from experience.
\newblock \emph{arXiv preprint arXiv:2508.04700}, 2025.

\bibitem[Tarasov et~al.(2023)Tarasov, Nikulin, Akimov, Kurenkov, and Kolesnikov]{tarasov2023corl}
Denis Tarasov, Alexander Nikulin, Dmitry Akimov, Vladislav Kurenkov, and Sergey Kolesnikov.
\newblock Corl: Research-oriented deep offline reinforcement learning library.
\newblock \emph{Advances in Neural Information Processing Systems}, 36:\penalty0 30997--31020, 2023.

\bibitem[Wang et~al.(2025)Wang, Qu, Huang, Chu, Lin, and Chen]{wang2025vl}
Haozhe Wang, Chao Qu, Zuming Huang, Wei Chu, Fangzhen Lin, and Wenhu Chen.
\newblock Vl-rethinker: Incentivizing self-reflection of vision-language models with reinforcement learning.
\newblock \emph{arXiv preprint arXiv:2504.08837}, 2025.

\bibitem[Wang et~al.(2024)Wang, Pan, Shi, Lu, Ren, Zhou, Zhan, and Li]{wang2024measuring}
Ke~Wang, Junting Pan, Weikang Shi, Zimu Lu, Houxing Ren, Aojun Zhou, Mingjie Zhan, and Hongsheng Li.
\newblock Measuring multimodal mathematical reasoning with math-vision dataset.
\newblock \emph{Advances in Neural Information Processing Systems}, 37:\penalty0 95095--95169, 2024.

\bibitem[Xia et~al.(2025)Xia, Zang, Gao, Li, and Zhou]{xia2025visionary}
Jiaer Xia, Yuhang Zang, Peng Gao, Yixuan Li, and Kaiyang Zhou.
\newblock Visionary-r1: Mitigating shortcuts in visual reasoning with reinforcement learning.
\newblock \emph{arXiv preprint arXiv:2505.14677}, 2025.

\bibitem[Xin et~al.(2025)Xin, Wang, Pan, Wang, Meng, Diao, Pi, and Zhang]{georeasoning}
Yue Xin, Wenyuan Wang, Rui Pan, Ruida Wang, Howard Meng, Shizhe Diao, Renjie Pi, and Tong Zhang.
\newblock Generalizable geometric image caption synthesis, 2025.
\newblock URL \url{https://arxiv.org/abs/2509.15217}.

\bibitem[Xing et~al.(2025)Xing, Dong, Zang, Cao, Liang, Huang, Wang, Wu, and Lin]{xing2025caprl}
Long Xing, Xiaoyi Dong, Yuhang Zang, Yuhang Cao, Jianze Liang, Qidong Huang, Jiaqi Wang, Feng Wu, and Dahua Lin.
\newblock Caprl: Stimulating dense image caption capabilities via reinforcement learning.
\newblock \emph{arXiv preprint arXiv:2509.22647}, 2025.

\bibitem[Yang et~al.(2025)Yang, Li, Yang, Zhang, Hui, Zheng, Yu, Gao, Huang, Lv, et~al.]{yang2025qwen3}
An~Yang, Anfeng Li, Baosong Yang, Beichen Zhang, Binyuan Hui, Bo~Zheng, Bowen Yu, Chang Gao, Chengen Huang, Chenxu Lv, et~al.
\newblock Qwen3 technical report.
\newblock \emph{arXiv preprint arXiv:2505.09388}, 2025.

\bibitem[Yarom et~al.(2023)Yarom, Bitton, Changpinyo, Aharoni, Herzig, Lang, Ofek, and Szpektor]{yarom2023you}
Michal Yarom, Yonatan Bitton, Soravit Changpinyo, Roee Aharoni, Jonathan Herzig, Oran Lang, Eran Ofek, and Idan Szpektor.
\newblock What you see is what you read? improving text-image alignment evaluation.
\newblock \emph{Advances in Neural Information Processing Systems}, 36:\penalty0 1601--1619, 2023.

\bibitem[Yu et~al.(2024)Yu, Baker, Chen, Ning, and Sun]{yu2024llasmol}
Botao Yu, Frazier~N Baker, Ziqi Chen, Xia Ning, and Huan Sun.
\newblock Llasmol: Advancing large language models for chemistry with a large-scale, comprehensive, high-quality instruction tuning dataset.
\newblock \emph{arXiv preprint arXiv:2402.09391}, 2024.

\bibitem[Yu et~al.(2025)Yu, Zhang, Zhu, Yuan, Zuo, Yue, Dai, Fan, Liu, Liu, et~al.]{yu2025dapo}
Qiying Yu, Zheng Zhang, Ruofei Zhu, Yufeng Yuan, Xiaochen Zuo, Yu~Yue, Weinan Dai, Tiantian Fan, Gaohong Liu, Lingjun Liu, et~al.
\newblock Dapo: An open-source llm reinforcement learning system at scale.
\newblock \emph{arXiv preprint arXiv:2503.14476}, 2025.

\bibitem[Yue et~al.(2024)Yue, Ni, Zhang, Zheng, Liu, Zhang, Stevens, Jiang, Ren, Sun, et~al.]{yue2024mmmu}
Xiang Yue, Yuansheng Ni, Kai Zhang, Tianyu Zheng, Ruoqi Liu, Ge~Zhang, Samuel Stevens, Dongfu Jiang, Weiming Ren, Yuxuan Sun, et~al.
\newblock Mmmu: A massive multi-discipline multimodal understanding and reasoning benchmark for expert agi.
\newblock In \emph{Proceedings of the IEEE/CVF Conference on Computer Vision and Pattern Recognition}, pages 9556--9567, 2024.

\bibitem[Zhang et~al.(2025)Zhang, Huang, Yao, Liu, Zhang, Lu, and Tao]{zhang2025r1}
Jingyi Zhang, Jiaxing Huang, Huanjin Yao, Shunyu Liu, Xikun Zhang, Shijian Lu, and Dacheng Tao.
\newblock R1-vl: Learning to reason with multimodal large language models via step-wise group relative policy optimization.
\newblock \emph{arXiv preprint arXiv:2503.12937}, 2025.

\bibitem[Zhang et~al.(2024)Zhang, Lu, Li, Ma, Chen, Xu, Luo, Zhang, Zhan, and Ye]{zhang2024wings}
Yi-Kai Zhang, Shiyin Lu, Yang Li, Yanqing Ma, Qing-Guo Chen, Zhao Xu, Weihua Luo, Kaifu Zhang, De-Chuan Zhan, and Han-Jia Ye.
\newblock Wings: Learning multimodal llms without text-only forgetting.
\newblock \emph{Advances in Neural Information Processing Systems}, 37:\penalty0 31828--31853, 2024.

\end{thebibliography}

\clearpage
\newpage
\beginappendix

\section{RLVR for VLMs}
\label{sec:grpo_appendix}

Our model training method is based on the Group Relative Policy Optimization (GRPO) algorithm~\cite{shao2024deepseekmath}. The core idea of GRPO is to sample multiple responses from the policy model for a given task and use verifiable rule-based rewards along with group relative advantages to replace the value model required in PPO, thereby reducing VRAM overhead.

Given a multimodal input $Q=\{i,q\}$, where $i$ denotes the image input and $q$ denotes the text question input; the policy model $\pi_{\theta}$ outputs a set of responses $\{o_1,...,o_G\}$; a rule-based reward function scores these responses, yielding a set of scalar rewards $\{r_1,...,r_G\}$.
The GRPO algorithm normalizes the group-relative rewards to obtain the response-level advantage:
\begin{equation}
 \hat{A}_i \;=\; \frac{r_i - \text{mean}(r_1,...,r_G)}{\text{std}(r_1,...,r_G)+\epsilon}
\label{eq:group_advantage}
\end{equation}
where $\epsilon$ is a small value used for numerical stability.

Similar to PPO~\cite{schulman2017proximal}, GRPO also employs a clipped objective on the importance weights and uses a KL divergence mechanism to maintain the stability of the training updates, preventing excessive deviation from the old policy distribution. The loss function is as follows:
\begin{align}
\begin{split}
&\mathcal{J}_{\text{GRPO}}(\theta) = \mathbb{E}\Bigg[
\frac{1}{G} \sum_{i=1}^G \frac{1}{|o_i|} \sum_{t=1}^{|o_i|}
\Bigg(
\min \Big( r_{i,t}(\theta) \hat{A}_{i}, \text{clip}\big( r_{i,t}(\theta), 1-\varepsilon_{l}, 1+\varepsilon_{h} \big) \hat{A}_{i} \Big)
- \beta D_{\text{KL}}(\pi_{\theta} \,\|\, \pi_{\text{ref}})
\Bigg)
\Bigg]
\end{split}
\label{eq:grpo_update}
\end{align}
where,
\begin{equation}
 r_{i,t}(\theta) = \frac{\pi_{\theta}(o_{i,t} \mid q, o_{i,<t})}{\pi_{\theta_{\text{old}}}(o_{i,t} \mid q, o_{i,<t})}.
\end{equation}
Here, we follow the work on DAPO~\cite{yu2025dapo} by decoupling the $\varepsilon$ values ($\varepsilon_{l}$ and $\varepsilon_{h}$), a method shown to help stabilize the entropy and enhance the model's exploration capability. The KL divergence loss, controlled by $\beta$, is used to prevent the model from deviating excessively from the reference policy distribution.

\section{Experiment Details}
\label{sec:experiment_appendix}

\subsection{Training Hyperparameter}
We use 8 x A100 GPUs for our experiments in total. GRPO baseline share the same hyperparameters with DoGe rl stage 2. Solver is not learnable, we only provide vllm sampling parameters. Table~\ref{tab:7b} and \ref{tab:3b} present the detailed experimental parameters for base models of 7B and 3B sizes, respectively.

\begin{table}[h!]
    \centering
    \begin{minipage}{0.48\textwidth}
        \centering
        \caption{3b-series experiment hyperparameters}
        \label{tab:3b}
        \begin{tabular}{llc}
            \toprule
            \textbf{Method} & \textbf{Parameter} & \textbf{Value} \\
            \midrule
            \multirow{2}{*}{Thinker} & train\_steps & 100 \\
            & train\_batch\_size & 64 \\
            & max\_prompt\_length & 768 \\
            & max\_response\_length & 4096 \\
            & lr & 1e-6 \\
            & n & 4 \\
            & temperature & 0.9 \\
            & clip\_ratio\_high & 0.24 \\
            & clip\_ratio\_low & 0.2 \\
            \midrule
            \multirow{2}{*}{Solver} & n & 4 \\
            & temperature & 0.9 \\
            \midrule
            \multirow{2}{*}{DoGe Stage 2*} & train\_steps & 150 \\
            & train\_batch\_size & 64 \\
            & max\_prompt\_length & 768 \\
            & max\_response\_length & 4096 \\
            & lr & 1e-6 \\
            & n & 8 \\
            & temperature & 1.0 \\
            & clip\_ratio\_high & 0.28 \\
            & clip\_ratio\_low & 0.2 \\
            \bottomrule
        \end{tabular}
    \end{minipage}
    \hfill  
    \begin{minipage}{0.48\textwidth}
        \centering
        \caption{7b-series experiment hyperparameters}
        \label{tab:7b}
        \begin{tabular}{llc}
            \toprule
            \textbf{Method} & \textbf{Parameter} & \textbf{Value} \\
            \midrule
            \multirow{2}{*}{Thinker} & train\_steps & 150 \\
            & train\_batch\_size & 48 \\
            & max\_prompt\_length & 768 \\
            & max\_response\_length & 4096 \\
            & lr & 1e-6 \\
            & n & 4 \\
            & temperature & 0.9 \\
            & clip\_ratio\_high & 0.24 \\
            & clip\_ratio\_low & 0.2 \\
            \midrule
            \multirow{2}{*}{Solver} & n & 4 \\
            & temperature & 0.9 \\
            \midrule
            \multirow{2}{*}{DoGe Stage 2*} & train\_steps & 150 \\
            & train\_batch\_size & 48 \\
            & max\_prompt\_length & 768 \\
            & max\_response\_length & 4096 \\
            & lr & 1e-6 \\
            & n & 8 \\
            & temperature & 1.0 \\
            & clip\_ratio\_high & 0.28 \\
            & clip\_ratio\_low & 0.2 \\
            \bottomrule
        \end{tabular}
    \end{minipage}
\end{table}

\subsection{Prompt Template for GRPO Training}
In this subsection, we present DoGe's prompts templates in RL stage 1 and stage 2, where \{information\} denotes the multimodal reasoning context with the question removed, \{Question\} denotes the corresponding reasoning question and \{deepthought\} denotes the analysis of Thinker$(\pi_T)$ to the corresponding question's contextual information.

\begin{tcolorbox}[
  colback=iris,   
  colframe=white,  
  title=\textbf{DoGe training stage 1},
  fonttitle=\bfseries,
  sharp corners
]

Thoroughly analyze the provided Contextual Artifacts and Design Information, from which the core question has been deliberately removed.\\
Your mission is to freely explore, deeply infer, and critically analyze the intrinsic structure, elemental relationships, and underlying constraints of this information.\\
Generate a detailed, insightful Deep-Thinking Report that will serve as the fundamental guiding knowledge for other models to efficiently solve the original, complete problem.\\

\#\#\# Informations \\
\{\textit{information}\}\\

You FIRST think about the reasoning process as an internal monologue and then provide the final answer. The reasoning process MUST BE enclosed within $\langle think \rangle$ $\langle /think \rangle$ tags. Your Deep-Thinking Report MUST follow $\langle /think \rangle$ tag.

\end{tcolorbox}

\begin{tcolorbox}[
  colback=iris,   
  colframe=white,  
  title=\textbf{Solver Prompt Template},
  fonttitle=\bfseries,
  sharp corners
]

\{\textit{Question}\} You FIRST think about the reasoning process as an internal monologue and then provide the final answer. The reasoning process MUST BE enclosed within $\langle think \rangle$ $\langle /think \rangle$ tags. The final answer MUST BE put in \textbackslash boxed\{\}. If the question is multiple-choice (single- or multi-select), put the final answer inside \textbackslash boxed\{\}, and format your answer as a Python list of uppercase letters in single quotes, separated by commas (e.g., \textbackslash boxed\{['D']\} or \textbackslash boxed\{['A','B']\}); otherwise, do not use a list.\\
To solve the problem above, you may refer to the expert analysis of the the given information and the problem scenario.\\
\#\#\# Expert Analysis:\\
{[Analysis Start]}\\
\{\textit{deepthought}\}\\
{[Analysis End]}
\end{tcolorbox}


\begin{tcolorbox}[
  colback=iris,   
  colframe=white,  
  title=\textbf{DoGe training stage 2},
  fonttitle=\bfseries,
  sharp corners
]

\{\textit{Question}\} You FIRST think about the reasoning process as an internal monologue and then provide the final answer. The reasoning process MUST BE enclosed within $\langle think \rangle$ $\langle /think \rangle$ tags. The final answer MUST BE put in \textbackslash boxed\{\}. If the question is multiple-choice (single- or multi-select), put the final answer inside \textbackslash boxed\{\}, and format your answer as a Python list of uppercase letters in single quotes, separated by commas (e.g., \textbackslash boxed\{['D']\} or \textbackslash boxed\{['A','B']\}); otherwise, do not use a list.
\end{tcolorbox}

\subsection{Prompt Template for Problem Synthesis}
For each type of disciplinary task, DoGe's reasoning problem synthesis framework offers two distinct paths: namely, synthesizing problems by sampling the Multimodal Knowledge Pool, and synthesizing variant problems by sampling the Seed Problem PoolS. We respectively denote these two paths as Synthesis 1 and Synthesis 2. \{rules\} part is optional, and it changes with the generation objective.
\begin{tcolorbox}[
  colback=iris,   
  colframe=white,  
  title=\textbf{Synthesis 1},
  fonttitle=\bfseries,
  sharp corners
]
You are a knowledgeable and skilled question designer. Design challenging problems based on the given informations and combining your own expertise as well.

The questions you design must follow these rules and format:

\{rules\}

Finally, present all questions and answers in a valid JSON block with the following format:\\

```json \\
{[\\
  \{"problem": "$\langle image \rangle$Question text here", "answer": "answer here", "images": {["image path"]}\},\\
  \{"problem": "Question text here", "answer": "answer here", "images": {[]}\}\\
]}\\
```\\

\#\# Informations\\
\{\textit{information}\}\\

Let's think step by step and put your final response in Json Block.

\end{tcolorbox}

\begin{tcolorbox}[
  colback=iris,   
  colframe=white,  
  title=\textbf{Synthesis 2},
  fonttitle=\bfseries,
  sharp corners
]
You are a knowledgeable and skilled question designer.
You will get some reference seed multi-modal problems. Your goal is to design challenging variant problems of given multimodal questions.

The questions you design must follow these rules and format:

\{rules\}

Finally, present all questions and answers in a valid JSON block with the following format:

```json \\
{[\\
  \{"problem": "$\langle image \rangle$Question text here", "answer": "answer here", "images": {["image path"]}\},\\
  \{"problem": "Question text here", "answer": "answer here", "images": {[]}\}\\
]}\\
```\\

\#\# Reference Seed Problems\\
\{examples\}\\

Let's think step by step and put your final response in Json Block.

\end{tcolorbox}

\subsection{Evaluation Details}

\begin{table}[h!]
    \centering
    \caption{Evaluation hyperparameters (Backbone Engine: \textbf{VLLM})}
    \label{tab:7b2}
    \begin{tabular}{llc}
        \toprule
        \textbf{Model Size} & \textbf{Parameter} & \textbf{Value} \\
        \midrule
        
        \multirow{2}{*}{7B} & tensor\_parallel\_size & 2 \\
        & gpu\_memory\_utilization & 0.85 \\
        & k(mean@k) & 4 \\
        & max\_tokens & 4096 \\
        & temperature & 0.7 \\
        \midrule
        
        \multirow{2}{*}{3B} & tensor\_parallel\_size & 1 \\
        & gpu\_memory\_utilization & 0.85 \\
        & k(mean@k) & 4 \\
        & max\_tokens & 4096 \\
        & temperature & 0.7 \\
        
        \bottomrule
    \end{tabular}
\end{table}

\section{Data Sample}

\subsection{Collected Raw Data Examples}
As shown in Section~\ref{subsubsec:data-preparation}, we have collected a large amount of unlabeled data, which serves as the foundation for the learning and evolution of self-evolving VLMs. In this section, we present examples of raw data from the three specific domains(multimodal math, chemistry, earth science).

\begin{tcolorbox}[
  colback=iris,   
  colframe=white,  
  title=\textbf{Chemistry},
  fonttitle=\bfseries,
  sharp corners
]

\textbf{Images}
\begin{center}
\includegraphics[width=0.4\linewidth]{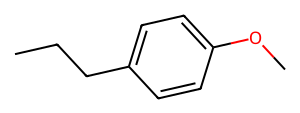}
\end{center}

\textbf{Raw Text Description}\\
\textbf{SMILES:} CCCC1=CC=C(C=C1)OC\\
\textbf{IUPAC:} 1-methoxy-4-propylbenzene\\
\textbf{Formula:} C10H14O, \textbf{MolWt:} 150.2210\\
\textbf{MolLogP:} 2.6477, \textbf{TPSA:} 9.2300\\
\textbf{HBD/HBA:} 0/1, \textbf{RotatableBonds:} 3\\
\textbf{RingCount:} 1

\end{tcolorbox}

\begin{tcolorbox}[
  colback=iris,   
  colframe=white,  
  title=\textbf{Earth Science},
  fonttitle=\bfseries,
  sharp corners
]

\textbf{Images}
\begin{center}
\includegraphics[width=0.4\linewidth]{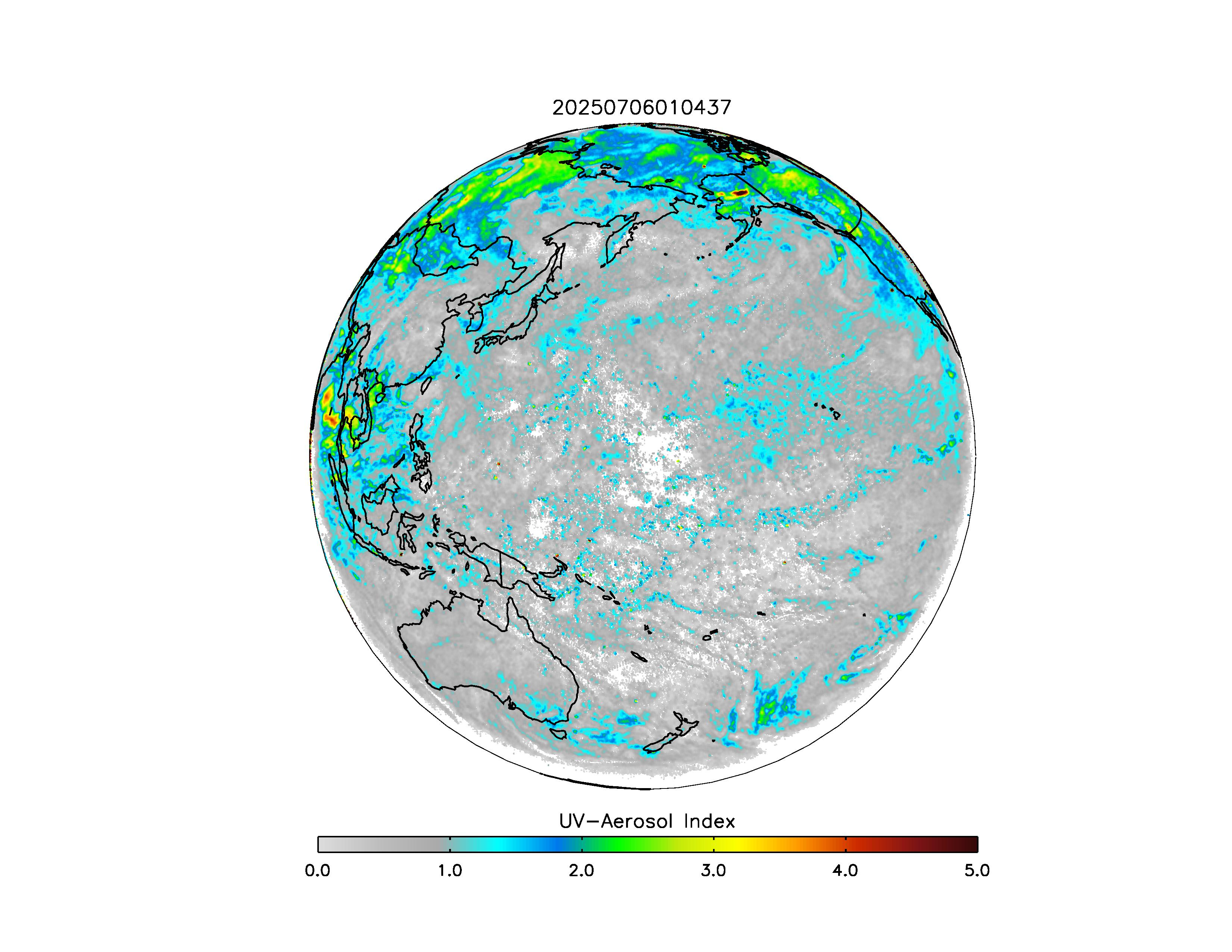}
\end{center}

\textbf{Raw Text Description}\\
Image Type: Aerosol; Images returned: 1 of 22 available\\
Image 1:Image saved at path1\\
Date: 2025-xx-xx\\
Caption: This image was taken by NASA's EPIC camera onboard the NOAA DSCOVR spacecraft

\end{tcolorbox}

\begin{tcolorbox}[
  colback=iris,   
  colframe=white,  
  title=\textbf{Math},
  fonttitle=\bfseries,
  sharp corners
]

\textbf{Images}
\begin{center}
\includegraphics[width=0.4\linewidth]{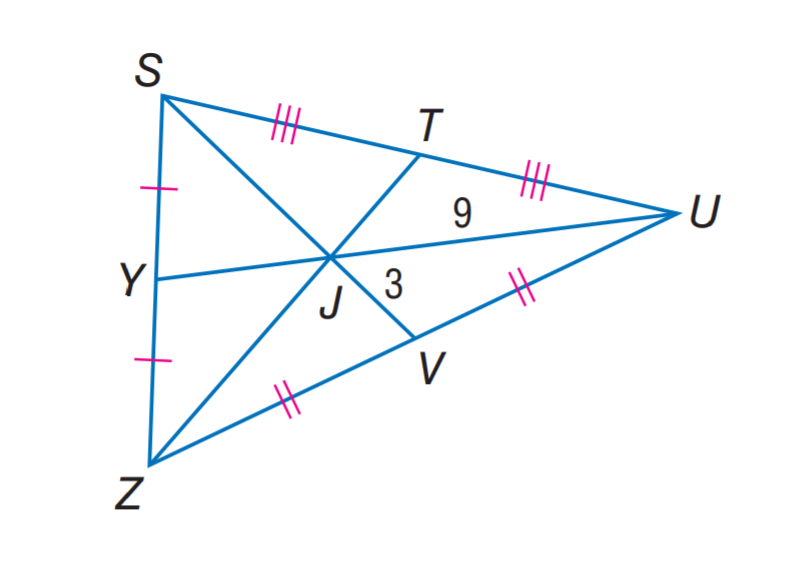}
\end{center}

\textbf{Raw Text Description}\\
$UJ=9, VJ=3$, and $ZT=18$.

\end{tcolorbox}

\subsection{Synthetic and Masked Problems}
We invoke the Gemini2.5-flash-lite to synthesize multimodal reasoning questions, and conduct masking processing on the original them. specifically, the questions are removed while only the contextual content is retained. We provide some examples below.

\begin{tcolorbox}[
  colback=iris,   
  colframe=white,  
  title=\textbf{Earth},
  fonttitle=\bfseries,
  sharp corners
]

\textbf{Image}:
\begin{center}
\includegraphics[width=0.4\linewidth]{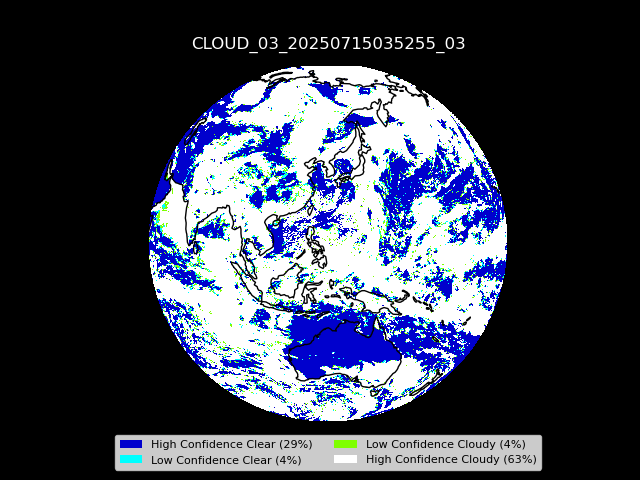}
\end{center}

\textbf{Question}:\\
Consider the cloud classification map. What might be the meteorological significance or interpretation of areas classified as `Low Confidence Clear` (cyan) or `Low Confidence Cloudy` (lime green) compared to `High Confidence Clear` (blue) and `High Confidence Cloudy` (white)?\\
A. `Low Confidence` areas typically indicate the presence of very stable atmospheric layers.\\
B. `Low Confidence Cloudy` might suggest thin clouds, haze, or cloud edges, which are harder to classify definitively.\\
C. `Low Confidence Clear` could indicate areas with thin cloud remnants, high humidity haze, or artifacts that don't strongly trigger cloud detection.\\
D. These low confidence zones are always associated with immediate precipitation events.\\
E. `High Confidence` classifications indicate areas where the atmospheric conditions are least dynamic.\\
F. The low confidence regions often represent transitional zones where atmospheric conditions are changing rapidly.\\
G. There is no significant difference in meteorological interpretation between high and low confidence areas.\\

\textbf{Answer}: B, C, F\\

\textbf{Masked Question Contextual Information}:\\
cloud classification confidence meaning
\end{tcolorbox}

\begin{tcolorbox}[
  colback=iris,   
  colframe=white,  
  title=\textbf{Chemistry},
  fonttitle=\bfseries,
  sharp corners
]

\textbf{Image}:
\begin{center}
\includegraphics[width=0.4\linewidth]{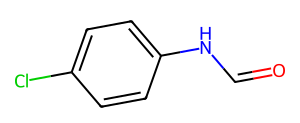}
\end{center}

\textbf{Question}:\\
Consider the molecule shown above. What is the sum of its Ring Count and the number of Hydrogen Bond Donors (HBD)?\\
A) 1\\
B) 2\\
C) 3\\
D) 4\\
E) 5\\
F) 6\\
G) 7

\textbf{Answer}: B\\

\textbf{Masked Question Contextual Information}:\\
Consider the molecule shown above.

\end{tcolorbox}

\begin{tcolorbox}[
  colback=iris,   
  colframe=white,  
  title=\textbf{Math},
  fonttitle=\bfseries,
  sharp corners
]

\textbf{Image}:
\begin{center}
\includegraphics[width=0.3\linewidth]{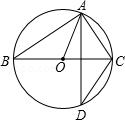}
\end{center}

\textbf{Question}:\\
In the given figure, BC is the diameter and O is the center of the circle. If $\angle$ADC = $32^\circ$, determine the measure of the central angle $\angle$AOB.\\

\textbf{Answer}: 116\\

\textbf{Masked Question Contextual Information}:\\
In the given figure, BC is the diameter and O is the center of the circle. If $\angle$ADC = $32^\circ$

\end{tcolorbox}

\section{Case Study}
In this section, we compare the output cases of DoGe, the base model, and the baseline to conduct an in-depth analysis of DoGe's role.

\begin{tcolorbox}[
  colback=iris,   
  colframe=chisa,  
  title=\textbf{Example Problem from MathVision},
  fonttitle=\bfseries,
  sharp corners
]

\textbf{Image}:
\begin{center}
\includegraphics[width=0.2\linewidth]{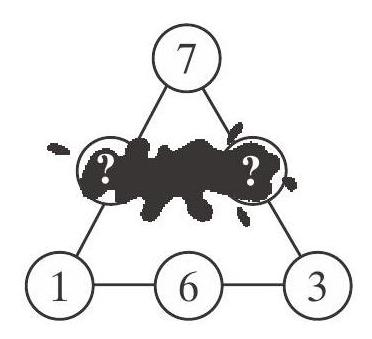}
\end{center}

\textbf{Question}:\\
The sums of the all the three numbers on each side of the triangle are equal. Two numbers happened to be stained with ink. How much is the sum of these two numbers?\\

\textbf{Answer}: 2

\end{tcolorbox}

\begin{tcolorbox}[
  colback=iris,   
  colframe=chisa,  
  title=\textbf{Baseline-7B Response.},
  fonttitle=\bfseries,
  sharp corners
]
\textbf{Avg Score: 0.0 (mean@4)}\\

\textbf{Response}:
$\langle think \rangle$
To solve this, we need to use the information that the sums of the numbers on each side of the triangle are equal. Let's denote the unknown numbers by \(x\) and \(y\), where \(x\) is the number on the left side (top of the left side), and \(y\) is the number on the right side (top of the right side).

\textcolor{red}{The sums of the numbers on the three sides of the triangle are:\\
1. \(1 + 6 + 7 = 14\)
2. \(7 + 3 + y = 14\)\\
3. \(x + 6 + 3 = 14\)}

From the second equation, we can solve for \(y\):
\[7 + 3 + y = 14\]
\[10 + y = 14\]
\[y = 4\]

From the third equation, we can solve for \(x\):
\[x + 6 + 3 = 14\]
\[x + 9 = 14\]
\[x = 5\]

Now, we need to find the sum of \(x\) and \(y\):
\[x + y = 5 + 4 = 9\]

Thus, the sum of the two numbers \(x\) and \(y\) is:
$\langle /think \rangle$
$\boxed{9}$

\end{tcolorbox}

\begin{tcolorbox}[
  colback=iris,   
  colframe=chisa,  
  title=\textbf{DoGe-7B Response.},
  fonttitle=\bfseries,
  sharp corners
]
\textbf{Avg Score: 0.5 (mean@4)}\\

\textbf{Response}:
$\langle think \rangle$
To solve this problem, we start by using the given condition that the sums of the numbers on each side of the triangle are equal. Let's denote the numbers at each of the three points as follows: the top point as \(7\), the left point as \(1\), the right point as \(3\), and the unknown numbers at the bottom as \(x\) and \(y\).\\

\textcolor{green}{1. The sum of the numbers on the left side is: \(1 + 6 + x\).\\
3. The sum of the numbers on the top side is: \(7 + y + x\).}\\

Since the sums are equal, we can set these expressions equal to each other:

\[
1 + 6 + x = 3 + 6 + y = 7 + y + x
\]

From the first two sides, we can set them equal and solve for \(x\) and \(y\):

\[
1 + 6 + x = 3 + 6 + y \implies 7 + x = 9 + y
\]

\[
\implies x - y = 2
\]

From the second and third sides, we can set them equal and solve for \(x\) and \(y\):

\[
3 + 6 + y = 7 + y + x \implies 9 + y = 7 + y + x
\]

\[
\implies 9 = 7 + x \implies x = 2
\]
Now, substituting \(x = 2\) back into the first equation:

\[
2 - y = 2 \implies y = 0
\]

So, the numbers are \(x = 2\) and \(y = 0\). The sum of these two numbers is:

\[
x + y = 2 + 0 = 2
\]
$\langle /think \rangle$
The sum of the two stained numbers is $\boxed{2}$.

\end{tcolorbox}

\begin{tcolorbox}[
  colback=iris,   
  colframe=chisa,  
  title=\textbf{Qwen2.5VL-7B Response.},
  fonttitle=\bfseries,
  sharp corners
]
\textbf{Avg Score: 0.0 (mean@4)}\\

\textbf{Response}:
To solve the problem, we need to determine the values of the two numbers that are stained with ink in the triangle. The sums of the numbers on each side of the triangle must be equal.

\textcolor{red}{Let's denote the numbers as follows:
- Top of the triangle: 7
- Left side: 1, 6, and x (unknown number)
- Right side: 3, y (unknown number), and z (unknown number)}

The sums of the numbers on each side of the triangle are equal. Therefore, we can set up the following equations:
\[ x - z = -3 \]
\[ z = x + 3 \]

Now, we need to find the value of \( y \) and \( z \) in terms of \( x \):
\[ y = x + 3 \]
\[ z = x + 3 \]

Since the sums of the numbers on each side must be equal, we can check the sum of the left side:
\[ 7 + 6 + x = 13 + x \]

And the sum of the right side:
\[ 7 + 3 + z = 10 + z \]

Since \( z = x + 3 \):
\[ 10 + (x + 3) = 13 + x \]

This equation is always true, so the values are consistent. Now, we need to find the sum of the two numbers that are stained with ink, which are \( y \) and \( z \):
\[ y + z = (x + 3) + (x + 3) = 2x + 6 \]

To find \( x \), we can use the sum of the numbers on the left side:
\[ 7 + 6 + x = 13 + x \]

Since the sums are equal, we can choose any consistent value for \( x \). Let's assume \( x = 0 \):
\[ y = 0 + 3 = 3 \]
\[ z = 0 + 3 = 3 \]

So, the sum of \( y \) and \( z \) is:
\[ y + z = 3 + 3 = 6 \]

Therefore, the sum of the two numbers that are stained with ink is:
\[
\boxed{6}
\]

\end{tcolorbox}

\end{document}